\documentclass[lettersize,journal]{IEEEtran}
\usepackage{amsmath,amsfonts}
\usepackage[linesnumbered,ruled,vlined]{algorithm2e}
\usepackage{array}
\usepackage{textcomp}
\usepackage{stfloats}
\usepackage{url}
\usepackage{verbatim}
\usepackage{graphicx}
\usepackage{cite}
\usepackage{xcolor}
\usepackage{diagbox}
\usepackage[justification=centering]{caption}
\usepackage{CJKutf8}
\usepackage{mathrsfs}
\usepackage{amssymb}
\usepackage{amsthm}
\usepackage{multirow}
\usepackage{pdfpages}
\usepackage{makecell}
\allowdisplaybreaks 
\makeatletter
\let\c@lofdepth\relax
\let\c@lotdepth\relax
\makeatother
\usepackage{subfigure}   
\usepackage{multicol}
\usepackage{fancyhdr}
\usepackage{graphicx} 
\usepackage{subcaption}

\newtheorem{thm}{\hspace{1em} Theorem}
\newtheorem{lem}{\hspace{1em} Lemma}
\newtheorem{assumption}{\hspace{1em} Assumption}

\begin{document}

\title{Federated Learning for Diffusion Models}
\author{Zihao Peng, Xijun Wang,~\IEEEmembership{Member,~IEEE,}
       Shengbo Chen*,~\IEEEmembership{Member,~IEEE,}
        Hong Rao*, Cong~Shen,~\IEEEmembership{Senior~Member,~IEEE} 
\thanks{
This work was supported in part by the Natural Science Foundation of Henan Province under Grant No.242300421402, in part by the High-level Overseas Talent Project of Jiangxi Province under Grant No.20232BCJ25026, in part by the Ji' An "Revealed Challenge" Key Core Common Technology List Project under Grant No.2022-1, and in part by the Ji' An Science and Technology Plan Special Project under Grant No.20222-151704 and Grant No.20222-151746, and in part by the National Natural Science Foundation of China under Grant No. 62271513. \textit{(Corresponding authors: Shengbo Chen (e-mail:ccb02kingdom@gmail.com); Hong Rao (e-mail:raohong@ncu.edu.cn).)}

Zihao Peng is with the School of Mathematics and Computer Science, Nanchang University, Nanchang 330000, China. 

Xijun Wang is with the School of Electronics and Information Technology, Sun Yat-sen University, Guangzhou 510006, China.

Shengbo Chen is with the School of Software, Nanchang University, Nanchang 330000, China.

Hong Rao is with the School of Software, Nanchang University, Nanchang 330000, China.

Cong Shen is with the Charles L. Brown Department of Electrical and Computer Engineering, University of Virginia, Charlottesville, VA 22904, USA.

}
}
\maketitle
\cfoot{\quad}

\begin{abstract}

Diffusion models are powerful generative models that can produce highly realistic samples for various tasks. Typically, these models are constructed using centralized, independently and identically distributed (IID) training data. However, in practical scenarios, data is often distributed across multiple clients and frequently manifests non-IID characteristics. Federated Learning (FL) can leverage this distributed data to train diffusion models, but the performance of existing FL methods is unsatisfactory in non-IID scenarios. To address this, we propose \textsc{FedDDPM}—Federated Learning with Denoising Diffusion Probabilistic Models, which leverages the data generative capability of diffusion models to facilitate model training. In particular, the server uses well-trained local diffusion models uploaded by each client before FL training to generate auxiliary data that can approximately represent the global data distribution. Following each round of model aggregation, the server further optimizes the global model using the auxiliary dataset to alleviate the impact of heterogeneous data on model performance. We provide a rigorous convergence analysis of \textsc{FedDDPM} and propose an enhanced algorithm, \textsc{FedDDPM+}, to reduce training overheads. \textsc{FedDDPM+} detects instances of slow model learning and performs a one-shot correction using the auxiliary dataset. Experimental results validate that our proposed algorithms outperform the state-of-the-art FL algorithms on the MNIST, CIFAR10 and CIFAR100 datasets.

\end{abstract}
\section{Introduction}
\label{Sec:1}

\begin{figure}
\centering
\includegraphics[scale=0.25]{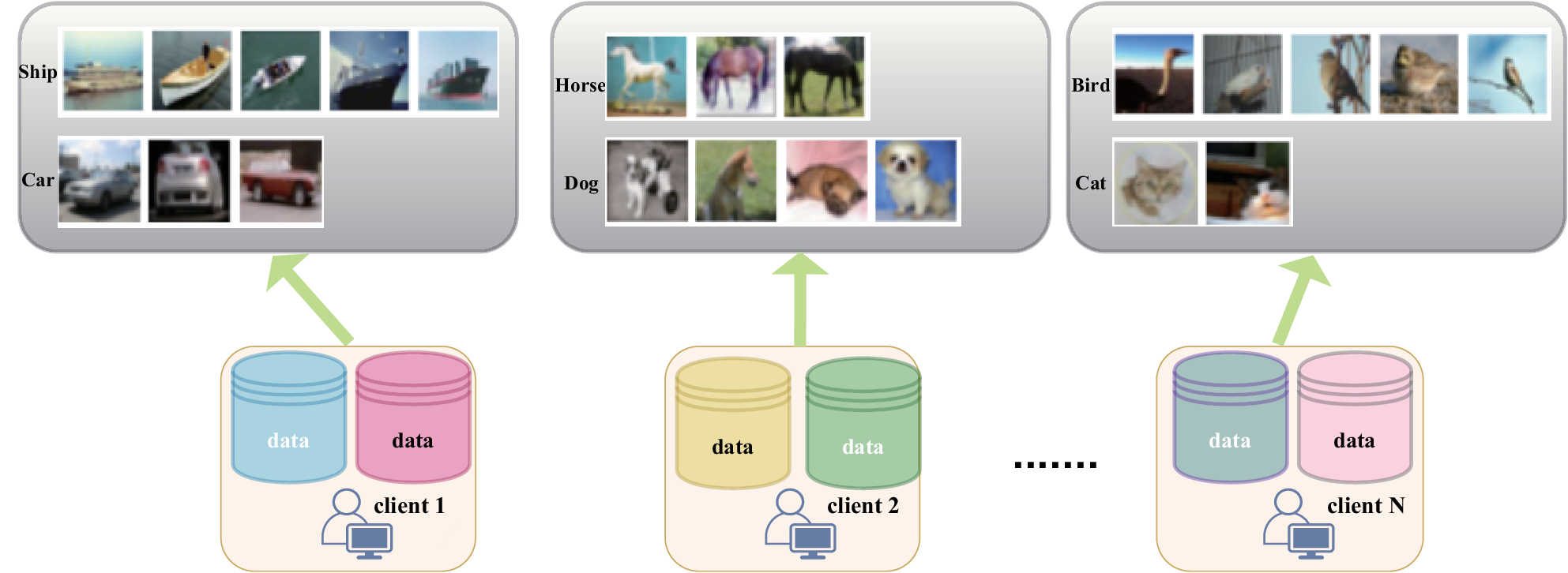}
\caption{Non-IID data distributions across clients.}
\label{fig1}
\end{figure}

\IEEEPARstart{D}{iffusion} models \cite{ho2020denoising} represent a powerful class of deep generative models acclaimed for their remarkable performance across diverse generative tasks, prominently in image synthesis \cite{yang2024improving}, sequence modeling \cite{ma2024plug}, and audio processing\cite{luo2024diff}. However, their success heavily relies on access to extensive centralized datasets \cite{nichol2021improved,song2020score}, a luxury often unattainable in real-world scenarios. Particularly in sensitive domains like healthcare \cite{chen2024channel}, data tends to be fragmented across disparate client systems, rendering centralized training impractical due to privacy concerns and data access restrictions. Moreover, the intrinsic non-IID nature of such decentralized data, stemming from diverse local environments and usage patterns, exacerbates the difficulty of training diffusion models\cite{qu2022rethinking, gao2022new}, as shown in Fig. \ref{fig1}.

Federated learning (FL)\cite{mcmahan2017communication}  provides a promising approach to enabling collaborative model training across distributed clients without the need for data sharing. This decentralized method effectively addresses data access limitations. Nonetheless, FL encounters significant challenges when handling non-IID data, as inconsistencies between global and local objectives can lead to biased model updates and degraded performance \cite{ye2023heterogeneous}. Although several enhanced FL techniques have been proposed to mitigate data heterogeneity, these approaches often do not fully utilize the unique capabilities of generative models, such as their ability to synthesize auxiliary data to enhance FL outcomes. Therefore, there remains substantial potential to improve the quality of samples generated by diffusion models within the conventional FL framework. It is of great interest to develop targeted solutions for diffusion models on non-IID data within the FL framework.

In this paper, we introduce \textsc{FedDDPM}—Federated Learning with Denoising Diffusion Probabilistic Models—a novel FL paradigm designed to address the challenges of training diffusion models on non-IID data. The proposed method harnesses the intrinsic data-generative capabilities of diffusion models to facilitate model training. Initially, each client trains a local diffusion model using its own data and then uploads this model to the server before FL begins. The server employs these models to generate auxiliary data through a proportional sampling strategy, merging the results to create a dataset that closely approximates the global data distribution. After each round of model aggregation, the global model is further refined using this auxiliary dataset to reduce update bias and improve convergence to the global optimum. Given that diffusion models are generally trained on deep neural networks with non-convex loss functions, we provide a convergence analysis of \textsc{FedDDPM} in the non-convex setting, theoretically demonstrating its eventual convergence. Additionally, we recognize that this collaborative mode requires serial training by the server after FL aggregation, potentially impacting the overall learning efficiency. To address this issue, we propose an enhanced algorithm, \textsc{FedDDPM+}, which uses auxiliary data for one-shot corrections when it detects slow learning progress in the global model. Specifically, we periodically evaluate the global model's performance during training to monitor its learning progress. If there are minimal differences between the current and previous evaluations, it indicates a slow learning pace. Experimental results demonstrate that \textsc{FedDDPM+} achieves superior performance without compromising FL efficiency.

We summarize the main contributions of this paper as follows.
\begin{enumerate}
\item 
We propose \textsc{FedDDPM}, a simple yet effective algorithm for training diffusion models via FL. It corrects biased learning through server refinement using the synthetic auxiliary dataset that reflects the global distribution. Additionally, we provide rigorous convergence analysis in non-IID scenarios.
\item
We introduce \textsc{FedDDPM+}, an enhanced algorithm that performs a one-shot correction using the auxiliary dataset. This approach greatly reduces the training overhead while maintaining comparable performance.
\item
We validate \textsc{FedDDPM} and the enhanced scheme \textsc{FedDDPM+} on the MNIST, CIFAR10 and CIFAR100 datasets with varying degrees of non-IID distributions. The simulation results show that our proposed algorithm outperforms state-of-the-art schemes.
\end{enumerate}

The remainder of this paper is organized as follows. Section \ref{Sec:2} provides an overview of related work. The main contribution of this paper, the \textsc{FedDDPM} algorithm, is presented in detail in Section \ref{Sec:3} and rigorously analyzed in Section \ref{Sec:4}. The enhanced \textsc{FedDDPM+} algorithm is presented in Section \ref{Sec:5}. Numerical experiment results are reported in Section \ref{Sec:6}. Finally, Section \ref{Sec:7} concludes the paper.

\section{Related work}
\label{Sec:2}

Diffusion models have gained popularity in recent years, and their combination with federated learning has become a promising research area\cite{li2024feddiff}. In this section, we review the relevant literature and discuss the potential of federated generative models as well as the challenges that need to be addressed.

\subsection{Diffusion Models}
Diffusion models, a novel state-of-the-art family of generative models, have recently gained significant attention in the deep learning community. Owing to the robust theoretical foundation and highly concise loss function design\cite{wang2023diffusion},  they can produce high-quality data in various applications \cite{croitoru2023diffusion,yang2023diffusion}. 
In particular, diffusion models utilize a Markov process to transform the original data distribution into random noise, and then inversely generate new samples that match the original data distribution.
Depending on the type of Markov process used, diffusion models can be classified into three major categories: denoising diffusion probabilistic models (DDPMs)\cite{ho2020denoising}, score-based generative models (SGMs)\cite{lee2023convergence}, and stochastic differential equations (SDEs)\cite{xue2024sa}. 
The iterative nature of Markov processes enables diffusion models to capture complex dependencies and correlations in the data, making them highly flexible and able to learn well over a wide range of data distributions.

\subsection{Federated Learning }

FL is a distributed machine learning paradigm that can harness large amounts of cross-silo data while preserving data privacy \cite{peng2023performance,10195234}. Nonetheless, FL encounters a significant obstacle arising from the heterogeneity of data across diverse clients, thereby adversely affecting its performance \cite{wu2022node,yangachieving,9790886}. The primary contributor to this performance decline lies in the inconsistency between the local objectives of clients and the global objective, resulting in a drift in the local updates \cite{panchal2023flash,gao2022feddc,jothimurugesan2023federated}. Subsequent to multiple local iterations, the aggregated model may deviate considerably from the global optimum.

Several strategies have been proposed to tackle the challenge of data heterogeneity in FL. In particular, researchers in \cite{sun2023dynamic,shi2023understanding} have employed proximal regularization in local training to constrain local updates. Despite mitigating the update magnitudes, these techniques fail to rectify the direction, thereby impeding convergence. On the other hand, approaches presented in \cite{karimireddy2020scaffold, gao2022feddc,sun2022fedspeed} undertake personalized local training by introducing additional control variables to approximate the drift gap. Nevertheless, these control variables heavily rely on historical update information, which may occasionally be inaccurate and fail to entirely eliminate the impact of non-IID data. The authors of \cite{qu2022generalized,fan2024locally} developed \textsc{FedSAM} based on the Sharpness-Aware Minimization (SAM) optimizer, making the global model less likely to fall into sharp valleys and enhancing generalization capabilities. However, it requires computing gradients twice in a single update step. Additionally, some studies aim to reduce bias from non-IID data in FL by carefully choosing a subset of clients to participate in training. The authors of \cite{mao2024joint} employ reinforcement learning to optimize model validation accuracy while penalizing increased communication rounds. Although this boosts accuracy, it comes at the cost of higher computational overhead. In contrast, CSFedAvg \cite{zhang2021client} prioritizes clients with lower non-IID degrees for faster convergence but may lead to some clients infrequently participating, affecting the model's generalization.

Some recent methods have attempted to address non-IID challenges rely on large external public datasets \cite{ gu2022fedaux, yang2024fedfed, yang2023fedvae,yang2022convergence}. Nevertheless, the practical acquisition of such public datasets presents considerable difficulties. To overcome this limitation, a few works have proposed performing knowledge distillation (KD) in a data-free manner by synthesizing data using generative models to improve the performance of classification tasks \cite{zhang2022fine, zhu2021data}. They propose a lightweight generator trained with the help of the representation power of a global model to generate class-related feature embedding data. In \cite{zhu2021data}, the synthesized data is employed to enhance local training for clients, while in \cite{zhang2022fine}, the synthesized data is used to refine the global model on the server. Unfortunately, the quality of generated data is closely tied to the performance of the global model. In cases of high data heterogeneity, this approach becomes questionable as it casts doubt on the quality of the generated synthetic data. Overall, existing methods still have some room for improvement in addressing the challenge of data heterogeneity, making it an area that requires further investigation.

\subsection{Integrating FL with Generative Models
}
Integrating FL with generative models holds great potential for overcoming data limitations and enhancing mutual benefits. By leveraging FL, generative models have made significant progress in areas such as Artificial Intelligence-Generated Content (AIGC) \cite{huang2024federated} and medical research \cite{guo2024diffusion}. At the same time, generative models contribute positively to the FL ecosystem, particularly in semi-supervised FL scenarios \cite{yang2023exploring, zhang2023gptfl}. Despite these advancements, research on effectively training generative models within the FL framework remains limited. Furthermore, collaborative training of generative models in real-world scenarios, such as inter-hospital collaborations, encounters challenges due to the non-IID nature of real-world data \cite{antunes2022federated}. This non-IID nature can compromise the generative capabilities, potentially affecting diagnostics and decision-making processes.

While some FL solutions attempt to handle non-IID data, they are generic approaches that often struggle to achieve satisfactory performance. As a result, customized FL methods have been developed specifically for Variational Autoencoders (VAEs) and Generative Adversarial Networks (GANs) to enhance their generative ability. For instance,  FedVAE \cite{polato2021federated}  uses a learning rate adaptation strategy to improve learning performance, and IFL-GAN \cite{li2022ifl} uses Maximum Mean Discrepancy (MMD) to adjust the aggregation weights of generators to promote global GAN learning. However, these methods lack rigorous convergence guarantees. In contrast, FedGAN \cite{rasouli2020fedgan} periodically averages the parameters of the generator and discriminator and provides a rigorous convergence analysis. However, it is a simple FL extension that does not fully exploit the data-generative capability of the generative model to deal with heterogeneous challenges. Other methods, such as MD-GAN \cite{hardy2019md}, UA-GAN \cite{zhang2021training} and Phoenix\cite{jothiraj2023phoenix}, use synthetic auxiliary data transferred by the server to train local discriminators to capture the overall distribution of data as much as possible. Nevertheless, this approach may lead to substantial transmission overhead and could pose risks of privacy leakage, making it impractical in certain scenarios. 

While the above-mentioned methods are valuable for joint generation, the most advanced diffusion models cannot benefit from them due to differences in their training processes. Specifically, GANs and VAEs require training a pair of networks, which complicates the training process when combined with FL. In contrast, diffusion models only need to be trained on a single deep neural network, making them well-suited for FL workflows. Recent studies have explored the integration of diffusion models with FL. For example, FedDiffuse \cite{de2024training} and FedDM \cite{tun2023federated} combine FedAvg with diffusion models for application in privacy-sensitive vision domains. They reduce communication overhead through the reuse of U-Net. FedTabDiff \cite{sattarov2024fedtabdiff} extends this approach to the generation of tabular data. Additionally, the authors of \cite{mendieta2024exploring} addresses performance degradation associated with differential privacy protection. FedDDA \cite{zhao2023federated} is proposed to solve the non-IID problem of classification tasks. FedDiff\cite{li2024feddiff} proposed a multimodal cooperative diffusion model framework and applied it to multimodal remote sensing data fusion. In this paper, we focus on tackling the issue of low-quality samples caused by distributed heterogeneous data and provide a rigorous convergence analysis. To the best of the authors' knowledge, this is an open question that has not been studied before.

\section{The \textsc{FedDDPM} Algorithm} 
\label{Sec:3}

In this section, we provide a detailed description of the proposed \textsc{FedDDPM} algorithm. Before delving into the specifics, we provide an overview of  preliminaries related to diffusion models and FL. 

\subsection{Preliminaries}
\subsubsection{Denoising Diffusion Probabilistic Models (DDPMs)}
DDPMs consist of two processes: a forward process that gradually adds Gaussian noise to data until it becomes fully Gaussian, and a reverse process where neural networks predict and remove noise to recover the original data.

\noindent\textbf{Forward process.}
Given a data distribution ${\xi}_0 \sim q({\xi}_0)$, we define a forward process $q$ that generates latents ${\xi}_1$ to ${\xi}_T$ by introducing Gaussian noise with variance $\beta_t \in (0, 1)$ at each time step $t$:
\begin{equation}
\label{eq:1}
q ( {\xi} _ { 1 } , \ldots , {\xi} _ { T } | {\xi} _ { 0 } ) : = \prod _ { t = 1 } ^ { T } q ( {\xi} _ { t } | {\xi} _ { t - 1 } ), \end{equation}

\begin{equation}
\label{eq:2}
{ q ( {\xi} _ { t } | {\xi} _ { t - 1 } ) : = \mathcal {N} ( {\xi} _ { t } ; \sqrt { 1 - \beta _ { t } } {\xi} _ { t - 1 } , \beta _ { t } I ) }. \end{equation}
With sufficient steps $T$ and a well-defined $\beta_t$ schedule, ${\xi}_T$ approaches an isotropic Gaussian distribution. More precisely, by defining $\alpha _ { t } : = 1 - \beta _ { t }$ and $\overline { \alpha } _ { t } : = \prod _ { s = 0 } ^ { t } \alpha _ { s }$, we have  $q ( {\xi} _ { t } | {\xi} _ { 0 } ) = \mathcal {N} ( {\xi} _ { t } ; \sqrt { \overline { \alpha } _ { t } } {\xi} _ { 0 } , ( 1 - \overline { \alpha } _ { t } ) I )$. To obtain a sample of ${\xi}_t$ given ${\xi}_0$, we can easily achieve this by sampling a Gaussian vector $\epsilon \sim \mathcal {N}(0,I)$ and then applying the following transformation:
\begin{equation}
\label{eq:3}
{\xi} _ { t } = \sqrt { \overline { \alpha } _ { t } } {\xi} _ { 0 } + \sqrt { 1 - \overline { \alpha } _ { t } } \epsilon.
\end{equation}
\textbf{Reverse process.}
The reverse process begins with a sample ${\xi}_T \sim \mathcal{N}(0, I)$ and aims to recover ${\xi}_0$. Since directly estimating $q({\xi}_{t-1}|{\xi}_t)$ is challenging, we approximate it using a neural network:
\begin{equation}
\label{eq:4}
p _ { w } ( {\xi} _ { t - 1 } | {\xi} _ { t } ) : = \mathcal {N} ( {\xi} _ { t - 1 } ; \mu _ { w } ( {\xi} _ { t } , t ) , \Sigma _ { w } ( {\xi} _ { t } , t ) ),
\end{equation}
where $w$ denotes the model parameters, and the mean and variance are modeled by the network. Starting from ${\xi}_T$, we iteratively sample from $p_w({\xi}_{t-1}|{\xi}_t)$ until reaching ${\xi}_0$. The training objective minimizes the loss to accurately simulate the reverse process:
\begin{align}
\label{eq:5}
l  =& - \operatorname { log } p _ { w } ( {\xi} _ { 0 } | {\xi} _ { 1 } ) + K L ( p ( {\xi} _ { T } | {\xi} _ { 0 } ) \| p ( {\xi} _ { T } ) ) \notag\\&  + \sum _ { t > 1 } K L ( p ( {\xi} _ { t - 1 } | {\xi} _ { t } , {\xi} _ { 0 } ) \| p _ { w } ( {\xi} _ { t - 1 } | {\xi} _ { t } ) ),      
\end{align}
where $KL$ refers to the Kullback-Leibler divergence.

\noindent\textbf{Training DDPMs.}
The authors of \cite{ho2020denoising} suggest adjusting the weights of different terms in $l$ to improve sample quality. The loss function $l$ is simplified as:
\begin{equation}
\label{eq:6}
l_ {simple} = E _ { t , {\xi} _ { 0 } , \epsilon } [ \| \epsilon - \epsilon_{ w } ( {\xi} _ { t } , t ) \| ^ { 2 } ],
\end{equation}
where ${\xi} _ { t }$ is computed from ${\xi}_0$ and $\epsilon$ by Eq.(\ref{eq:3}), and $\epsilon_w$ refers to the outputs of a deep neural network, usually with a U-Net-based\cite{ronneberger2015u} structure, that predicts the noise vector $\epsilon$ based on ${\xi}_t$ and $t$.

\subsubsection{Federated Learning}
For the federated learning system, we consider that client $i$ possesses a local dataset $\mathcal D_i$ consisting of $m_i$ samples intended for participation in FL training. The corresponding loss function for client $i$ is defined as $f_{i}(w) = \frac{1}{m_i}\sum_{\xi \in \mathcal{D}_i} \ell(w;\xi)$, where $w \in \mathbb{R}^d$ represents the machine learning model to be optimized, and $\ell(w;\xi)$ is the loss function evaluated at data sample $\xi$ with model $w$. Assuming there are a total of $N$ clients, the overall training objective is formulated as the weighted sum of individual client objectives:

\begin{equation}
\label{eq:00}
f(w) = \sum_{i=1}^{N} p_{i} f_{i}(w),
\end{equation}
where $p_{i} = \frac{m_{i}}{m}$ denotes the ratio of the local dataset size $m_{i}$ allocated to client $i$ to the global dataset size $m$.

The goal is to find the optimal solution to the overall objective function $f(w)$, denoted as $w^*$, which is defined as:

\begin{equation}
\label{eq:0}
w^* = \underset{w}{\operatorname{arg min}} f(w).
\end{equation}

\begin{figure}
\centering
\includegraphics[width=98mm]{{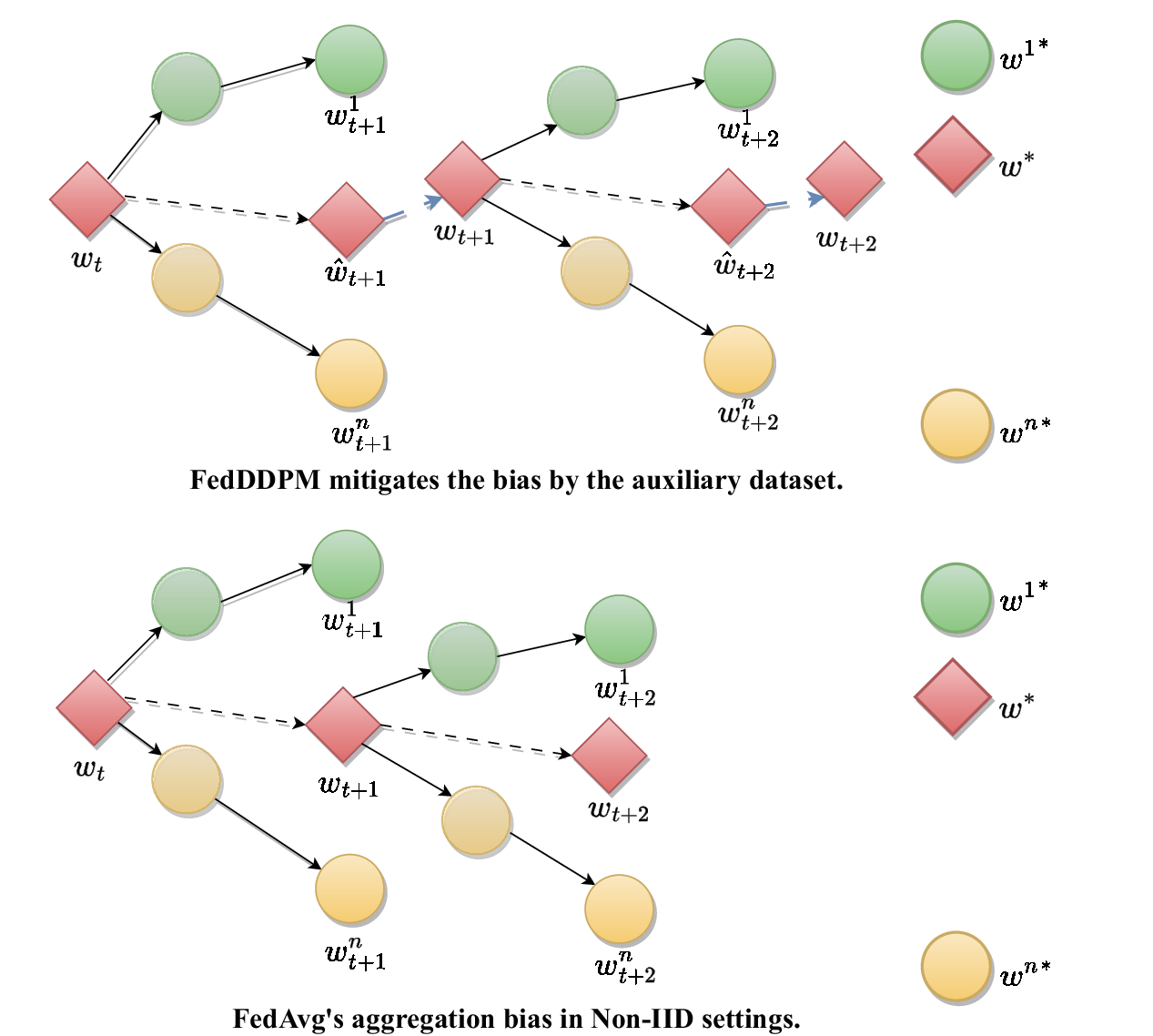}}
\caption{Illustration of model training differences between \textsc{FedAvg} and \textsc{FedDDPM}. }
\label{fig2}
\end{figure}
\begin{algorithm}[ht]
\caption{The \textsc{FedDDPM} Algorithm.}
\label{alg:algorithm1}
\KwIn{Training set $D$, Initial model $w_{0}$}
\KwOut{The trained model $w_{\text{final}}$}

Server collects well-trained local models \( \leftarrow \textbf{\textsc{Warmup}} \);
Create an auxiliary dataset \( \mathcal{A} = \sum_{i \in S} \text{Sampling}(w_{\text{warmup}}^i, |\mathcal{A}_i|) \);

\For{each round $t \in [T] \triangleq \{0, 1, \cdots, T-1 \}$}{
    $n \leftarrow \max(N\times p,1)$ \;
    $Z_t \leftarrow \text{(A random subset with $n$ clients)}$ \;
    
    \For{$i \in Z_t$ in parallel}{
        ${w}_{t+1}^i\leftarrow \textbf{Client Update}(w_{t},K)$;
    }
    $\hat{w}_{t+1}={w}_{t}+ \sum_{i\in Z_t} \frac{1}{n}({w}_{t+1}^i-{w}_{t})$ \;
    $\hat{w}_{t+1,0}=\hat{w}_{t+1}$;
    
    \For{$e = 0$ \KwTo $E-1$}{
        $\hat{w}_{t+1,e+1} = \hat{w}_{t+1,e} -  \eta_t \tilde{\nabla} f_\mathcal A (\hat{w}_{t+1,e})$; // Server uses auxiliary dataset $\mathcal A$ to correct direction
    }
    ${w}_{t+1} = \hat{w}_{t+1,E}$;
}

\SetKwProg{ClientUpdate}{Client Update}{}{}
\ClientUpdate{($w_t, K$)}{
    \( w_{t,0}^i \leftarrow w_t \);
    
    \For{each local epoch \( \tau \in [K] \triangleq \{0, 1, \cdots, K-1\} \)}{
        \( w_{t,\tau+1}^i = w_{t,\tau}^i - \zeta_t \tilde{\nabla} f_i(w_{t,\tau}^i) \);
    }
    Upload \( w_{t,K}^i \) to Server;
}

\end{algorithm}
\textsc{FedAvg}\cite{mcmahan2017communication} is proposed to solve the problem, and it works as follows. The server randomly selects a subset of clients and distributes the initial model to them. The clients then perform stochastic gradient descent (SGD) locally for several rounds before uploading their trained model to the server for aggregation. This process continues for $T$ training rounds, after which the server obtains the final global model. 

In this paper, we set the local loss function of the federated diffusion models to be consistent with Eq. (\ref{eq:6}). We clarify that this setting is one popular form of diffusion model families.  However, our proposed algorithm is not limited to this setting and can be applied to other types of diffusion models, such as SGMs  and SDEs.

\begin{figure*}
\centering
\includegraphics[width=\textwidth,height=10cm]{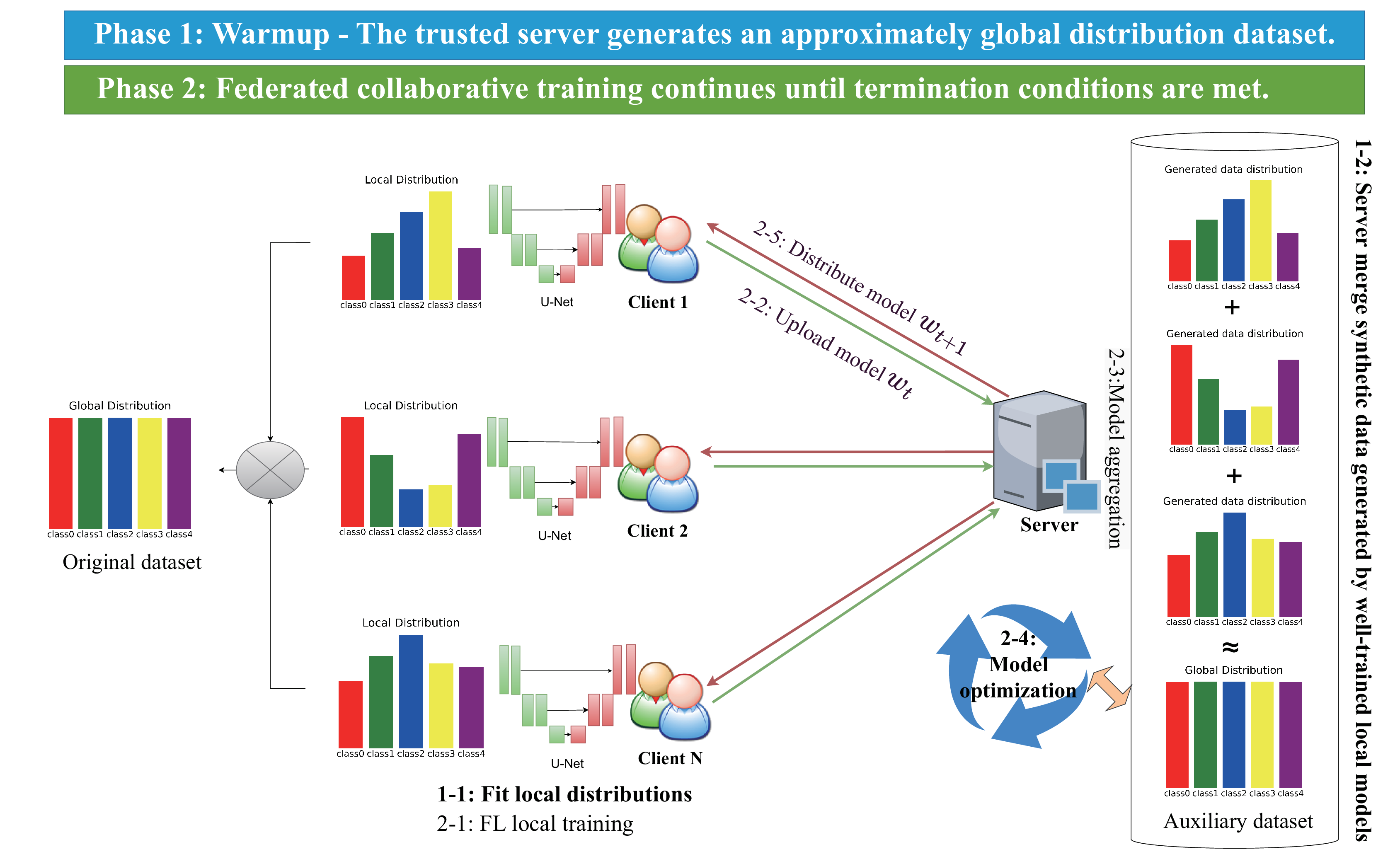}
\caption{The Workflow of \textsc{FedDDPM}.}
\label{fig3}
\end{figure*}
\subsection{Algorithm Description}
\label{subsec:des}
The procedure of the \textsc{FedDDPM} algorithm is formally presented in Algorithm \ref{alg:algorithm1}. Before the beginning of FL training, 
each client $i \in$ $S$, where $S$ is the set of all clients, conducts local training to fit its local data distribution as closely as possible, a process referred to as "\textbf{\textsc{Warmup}}".
Afterwards, the client $i$ uploads the well-trained model ${w}_{\textsc{warmup}}^i$ to a trusted server, which utilizes it to generate samples based on the model's learned distribution.\footnote{In this paper, we utilize the sampling algorithm proposed in \cite{ho2020denoising} to generate data. Although accelerated sampling algorithms like Dpm-Solver \cite{zheng2024dpm} are available, they may sacrifice the quality of the resulting image. Hence, we opt for the original non-accelerated sampling algorithm to ensure the production of high-quality images.} Once all clients' models have generated their respective data, the server aggregates these samples to construct an auxiliary dataset $\mathcal{A}$. To approximate the global data distribution, two key alignment steps are undertaken for the auxiliary dataset. First, the generated data is aligned with each client’s local data distribution, leveraging the diffusion model’s strong capability to accurately capture underlying data patterns. Second, the distribution of the generated data is matched to the global data distribution by ensuring that the number of samples produced by each client’s model is proportional to the size of its local dataset. This proportionality guarantees that each client's contribution to the auxiliary dataset accurately reflects its representation within the global dataset.

We assume that the distribution of the global dataset $D$ follows $\lambda$, i.e., each data sample $ \xi \in D$ is sampled from the distribution $\lambda$.  
For the auxiliary dataset $\mathcal A$, we also assume that its distribution obeys $\lambda$, which leads to training on the auxiliary dataset being consistent with solving the learning problem of Eq. (\ref{eq:00}).  
Consequently, the updates from the auxiliary dataset will converge towards the global optima. In \textsc{FedDDPM},  we use updates from the auxiliary dataset to compensate for the updates of the global model, alleviating global model bias caused by non-IID data. In contrast, the vanilla \textsc{FedAvg} algorithm only employs updates from local data to advance the model. The main training difference between \textsc{FedDDPM} and \textsc{FedAvg} is shown in Figure \ref{fig2}.

The detailed description of the algorithm is as follows:  In the FL training at round  $t \in [ T ] \triangleq \{ 0,1 , \cdots , T-1 \} $, a subset $Z_t$ of $n$ clients is randomly selected from all $N$ clients according to the client ratio $p$. For each client $i \in Z_t$, the server sends the latest  model $w_t$ to them to perform the client update. Then each client updates its model by running SGD locally $K$ steps and obtaining $w_{t+1}^i$.  Specifically, in the $\tau$-th iteration of client $i$'s local SGD, a data sample  $\xi _ { t , \tau } ^ { i }$  is randomly selected from the client's local dataset, and calculates the gradient $\tilde{\nabla}  {f_i}(w_{t,\tau}^i) = \nabla \ell ( w _ { t , \tau } ^ { i } ; \xi _ { t , \tau } ^ { i } )$. After the client performs $K$ step SGD with learning rate $\zeta_{t}$, they upload the resulting $w_{t,K}^i$ to the server, which receives the model and denotes as $w_{t+1}^i$. Once the server receives all local updates and aggregates to get the new model $\hat{w}_{t+1}$, that is 
\begin{equation}
\hat{w}_{t+1}={w}_{t}+ \sum_{i\in Z_t} \frac{1}{n}({w}_{t+1}^i-{w}_{t}).
\end{equation}
The server then uses this model as a new starting point and proceeds to advance the global model to ${w}_{t+1}$ by running the SGD $E$ steps on the auxiliary dataset $\mathcal{A}$. 
The core rationale behind the \textsc{FedDDPM} correction mechanism stems from its reliance on a secondary dataset, a more balanced and globally representative dataset. This mechanism plays a crucial role in supporting federated diffusion models by facilitating the alignment of global features through additional training on the server, thus improving its overall generation performance. Simultaneously, the server introduces its own server-side step size $\eta_t$ to govern the model updates. The entire workflow of \textsc{FedDDPM} is visually illustrated in Figure \ref{fig3}.

Notably, our proposed algorithm strategically leverages the generative model's characteristics to alleviate non-IID aggregation bias. This is achieved by integrating gradient updates from an auxiliary dataset directly on the trusted server. While some existing works explore server-side collaborative training architectures, they can't generate IID data on the server to compensate for the gradient bias, and instead rely on techniques such as historical gradients or global momentum \cite{10546478}. In addition, a key feature of our methodology is its emphasis on maintaining data privacy. Unlike conventional approaches such as \cite{hardy2019md, yoon2020fedmix}, which distribute synthetic datasets to individual clients for local training improvement, our strategy safeguards data privacy by refraining from granting clients access to synthetic auxiliary datasets mimicking other clients' data sources. In Section \ref{Sec:4}, we thoroughly study the convergence analysis of \textsc{FedDDPM}. To alleviate the overhead associated with additional training, Section \ref{Sec:5} presents \textsc{FedDDPM+}, a framework that optimizes training efficiency through one-shot correction.

\section{Convergence Analysis of \textsc{FedDDPM}}
\label{Sec:4}
In this section, we discuss the primary theoretical findings on the convergence behavior of \textsc{FedDDPM}. Our analysis is centered on the non-convex loss function since the training of diffusion models typically involves the use of deep neural networks based on the U-Net structure \cite{saharia2022image}. Without loss of generality, we consider scenarios in which a subset of clients is sampled without replacement.

For non-convex loss functions, the gradient descent algorithm will converge to a local minimum or saddle point. In such cases, it is common to use expected gradient norms as a criterion for convergence. Specifically, an algorithm is considered to have achieved an $\epsilon$-suboptimal solution if:
\begin{align}
\label{eq:10}
    \frac{1}{T} \sum_{t=0}^{T-1}\mathbb E \left\|\nabla f(w_t)\right\|^2   \leq \epsilon,
\end{align}
which guarantees the convergence to a stationary point \cite{wang2021cooperative,das2022faster}.
Moreover, $f$ is bounded below, i.e., for any vector $w$, $f^* := \inf_{w} f(w) > -\infty$ \cite{xin2021improved}. Next, we make the following commonly adopted assumptions about the loss function and model parameters to facilitate analysis.
\begin{assumption}\label{assumption:1}(L-Lipschitz Continuous Gradient)
The loss function $\ell$ is $L$-smooth with respect to $w$, i.e., for any $w,\hat{w}\in\mathbb{R}^d$ and $\xi \in\mathcal{D}$, we have
\begin{align}
\left\|\nabla\ell(w;\xi )-\nabla\ell(\hat{w};\xi )\right\|\leq L\left\|w-\hat{w}\right\|.
\end{align}
\end{assumption}
\begin{assumption}
\label{assumption:2} (Unbiased Local Stochastic Gradient) 
The local gradient estimator $\tilde{\nabla} f_i(w)$ is unbiased for ${\forall} i \in [N]$, i.e.,
\begin{align}
    \mathbb{E}[\tilde{\nabla} f_i(w)]=\nabla  f_i(w).
\end{align}
\end{assumption}
\begin{assumption}
\label{assumption:3}(Bounded Local and Global Variance) The variance of the local stochastic gradient is bounded, and the variance of the local and global gradients is also bounded.
\begin{align}
    \mathbb{E}\left\|\tilde{\nabla}  f_i(w)-{\nabla}f_i(w)\right\|^2\leq \sigma_l^2, \mathbb{E}\left\|\nabla f_i(w)- \nabla f(w)\right\|^2 \leq \sigma_g^2.
\end{align}
\end{assumption}

\begin{assumption}
\label{assumption:4}
(Global Unbiasedness of Auxiliary Dataset Gradients) In an ideal scenario, we assume that the global dataset and the auxiliary dataset share identical distributions. Under this gentle assumption, it is presumed that the gradient $\tilde{\nabla} f_\mathcal{A}(w)$ computed from the auxiliary dataset maintains global unbiasedness \cite{haddadpour2021federated, xiong2021privacy}.
\begin{align}
\mathbb{E}[\tilde{\nabla} f_\mathcal A(w)] = \nabla f(w).
\end{align}
\end{assumption}

Assumption \ref{assumption:1} implies that the gradient $\nabla\ell(w)$ satisfies the Lipschitz continuity condition. Consequently, both the local function $f_{i}$ and the global function $f$ have Lipschitz continuous gradients. Assumption \ref{assumption:2} assumes that the client's stochastic gradient is unbiased. 
Assumption \ref{assumption:3} indicates that the variance of the client's local gradient is bounded by $\sigma_l$, and the variance of the gradient between the local and global is bounded by $\sigma_g$. 
Assumption \ref{assumption:4} suggests that the gradient $\tilde{\nabla} f_\mathcal A(w)$, calculated using the auxiliary dataset, is globally unbiased. This means that its expected value is consistent with the gradient of the overall loss function $\nabla f(w) $. The main theorem we propose about the convergence of \textsc{FedDDPM} is as follows.
\begin{thm}
\label{theorem:1}
Suppose Assumptions \ref{assumption:1}, \ref{assumption:2} and \ref{assumption:3} hold, the stepsize $\eta_t$ is set to $\frac{1}{LE\sqrt{T}}$, and $\zeta_t$ is set to $\frac{1}{2LK\sqrt{T}}$, subject to the following lemma's constraints. The convergence of \textsc{FedDDPM} with non-convex loss functions satisfies
\begin{align}
&\frac{1}{T} \sum_{t=0}^{T-1}\mathbb{E}\left\|\nabla f(w_t)\right\|^2 \leq \frac{1}{C} \left(\frac{2L(f(w_0)-f^*)}{\sqrt{T}} \notag\right.\\&\left. +(\frac{3}{8KT}+\frac{9(N-n)}{16n(N-1)K T^\frac{3}{2}})(\sigma_l^2+6K\sigma_g^2) + \frac{\sigma_l^2}{4Kn\sqrt{T}} \notag\right.\\&\left. +\frac{(E-1)(2E-1)\sigma_l^2}{6E^2T}+
\frac{\sigma_l^2}{\sqrt{T}}+\frac{3(N-n)\sigma_g^2}{4n(N-1)\sqrt{T}}\right).
\end{align}
\end{thm}
\textbf{Proof:}
We initially introduce two lemmas. The proof of Theorem \ref{theorem:1} relies on the propositions established in these lemmas.

\begin{lem} \label{lem:1}
Under the conditions  
\(
\frac{1}{2} - 9K^2\zeta_t^2L^2 - \frac{\zeta_t L (N-n)}{2n(N-1)}\big(54K^3\zeta_t^2L^2 + 3K\big) > C \geq 0
\),
where \(C\) is a constant and \(\zeta_tKL \leq \frac{n(N-1)}{N(n-1)}\), the expected loss of the global model \(w_t\) and the aggregated model \(\hat{w}_{t+1}\) satisfies: 
\begin{align}
&\mathbb{E}f(\hat{w}_{t+1}) \leq\mathbb{E}f(w_t)-K\zeta_tC\mathbb{E}\left\|\nabla f(w_t)\right\|^2  + \frac{K\zeta_t^2L\sigma_l^2}{2n} \notag\\&+ (\frac{3}{2} K^2 \zeta_t^3 L^2+\frac{(N-n)}{2n(N-1)}9K^3L^3\zeta_t^4) (\sigma_l^2+6K\sigma_g^2) \notag\\& + \frac{(N-n)}{2n(N-1)}3K^2L\zeta_t^2\sigma_g^2. 
\end{align}
\end{lem}

\begin{lem} \label{lem:2}
The relationship between the expected loss value of the global model $w_{t+1}$ and the expected loss value of $\hat{w}_{t+1}$ corrected by the server is as follows:
\begin{align}
\mathbb{E}f(w_{t+1}) \leq& \mathbb E f(\hat{w}_{t+1})-\frac{\eta_tE}{2}\mathbb{E}\left\|\nabla f(\hat{w}_{t+1})\right\|^2+\frac{L\eta_t^2E^2\sigma_l^2}{2}\notag\\&+\frac{L^2\eta_t^3 E(E-1)(2E-1)\sigma_l^2}{12}.
\end{align}
\end{lem}
\textbf{Proof of Theorem \ref{theorem:1}}:
\noindent\textrm Now we have established the main building modules to prove Theorem \ref{theorem:1}. Next, we combine the results of Lemma \ref{lem:1}, \ref{lem:2} to start the proof. That is,
\begin{align}
\label{eq:kk}
\mathbb{E}&f(w_{t+1}) \leq \mathbb{E}f(w_t)-(K\zeta_tC\mathbb{E}\left\|\nabla f(w_t)\right\|^2 \notag\\ &+\frac{\eta_tE}{2}\left\|\nabla f(\hat{w}_{t+1})\right\|^2 )+ \frac{(N-n)}{2n(N-1)}3K^2L\zeta_t^2\sigma_g^2\notag\\&+(\frac{3}{2} K^2 \zeta_t^3 L^2 +\frac{(N-n)}{2n(N-1)}9K^3L^3\zeta_t^4) (\sigma_l^2+6K\sigma_g^2)\notag\\&+\frac{K\zeta_t^2L\sigma_l^2}{2n}+\frac{L\eta_t^2E^2\sigma_l^2}{2}+\frac{L^2\eta_t^3 E(E-1)(2E-1)\sigma_l^2}{12}.
\end{align}

We sum both sides of Eq.~(\ref{eq:kk}) over \( t = 0, \dots, T - 1 \), and rearrange the terms. Specifically, we replace \( \mathbb{E}f(w_T) \) with \( f^* \), which is the lower bound of the loss function $f$. This yields
\begin{align}
\sum_{t=0}^{T-1}&(K\zeta_tC\mathbb{E}\left\|\nabla f(w_t)\right\|^2 +\frac{\eta_tE}{2}\mathbb{E}\left\|\nabla f(\hat{w}_{t+1})\right\|^2)  
\notag\\
\leq& f(w_0)-f^*+(\frac{(N-n)}{2n(N-1)}9K^3L^3\sum_{t=0}^{T-1}\zeta_t^4\notag\\+&\frac{3}{2} K^2 L^2 \sum_{t=0}^{T-1}\zeta_t^3) \times  (\sigma_l^2+6K\sigma_g^2) + \frac{KL\sigma_l^2}{2n} \sum_{t=0}^{T-1}\zeta_t^2\notag\\+&\frac{LE^2\sigma_l^2}{2}\sum_{t=0}^{T-1}\eta_t^2 + \frac{(N-n)}{2n(N-1)}3K^2L\sigma_g^2 \sum_{t=0}^{T-1} \zeta_t^2\notag \\+&\frac{L^2E(E-1)(2E-1)\sigma_l^2}{12}\sum_{t=0}^{T-1}\eta_t^3.
\end{align}
We set $\eta_t=\frac{1}{LE\sqrt{T}}$, $\zeta_t=\frac{1}{2LK\sqrt{T}}$  which yields the desired result.

\textbf{Remark 1:} Our analysis indicates that the \textsc{FedDDPM} algorithm exhibits a convergence rate of $\mathcal{O}(\frac{1}{\sqrt{T}})$, which is consistent with the convergence rate of local SGD \cite{wang2021novel}. 

\textbf{Remark 2:} Upon examination of Eq. (\ref{eq:kk}), we observe that, unlike conventional FL algorithms, additional terms emerge with each round of gradient descent. These additional terms consist of $-\frac{\eta_tE}{2}\mathbb{E}\left\|\nabla f(\hat{w}_{t+1})\right\|^2$, representing the enhanced degree of gradient descent on the server, as well as $\frac{L\eta_t^2E^2\sigma_l^2}{2}$ and $\frac{L^2\eta_t^3 E(E-1)(2E-1)\sigma_l^2}{12}$, which originate  from the stochastic noise introduced by SGD.
When employing a suitably small learning rate $\eta_t$, these higher-order noise terms become negligible compared to the dominant influence of the first term\cite{allen2016variance, lei2021learning}. Consequently, the expected loss value of the global model decreases more rapidly in the early learning phase compared to algorithms without server-side correction.

\textbf{Remark 3:} It is noteworthy that comparing algorithm performance based on the convergence analysis of non-convex loss functions is complex. The primary challenge lies in the lack of assurance that the objective function in Eq. (\ref{eq:10}) will guide the model to a consistent optimum. As learning progresses, algorithms may converge to different stationary points, making a straightforward analytical comparison of their convergence behaviors difficult. In Section \ref{Sec:6}, we elucidate their performances through empirical experimentation.

\begin{algorithm}[ht]
\caption{The \textsc{FedDDPM+} Algorithm.}
\label{alg:algorithm2}
\KwIn{Training set $D$, Initial model $w_{0}$}
\KwOut{The trained model $w_{\text{final}}$}
Server collects well-trained local models \( \leftarrow \textbf{\textsc{Warmup}} \);
Create an auxiliary dataset \( \mathcal{A} = \sum_{i \in S} \text{Sampling}(w_{\textsc{warmup}}^i, |\mathcal{A}_i|) \);

\For{each round $t \in [T] \triangleq \{0, 1, \cdots, T-1 \}$}{
    $n \leftarrow \max(N\times p,1)$ \;
    $Z_t \leftarrow \text{(A random subset with $n$ clients)}$ \;
    
    \For{$i \in Z_t$ in parallel}{
        ${w}_{t+1}^i\leftarrow \textbf{Client Update}(w_{t},K)$;
    }
    $w_{t+1}={w}_{t}+ \sum_{i\in Z_t} \frac{1}{n}({w}_{t+1}^i-{w}_{t})$ \;
    
    \If{$t \mod 10 == 0$ \textbf{and} $\textsc{QuickTest}({w}_{t+1})$}{
        $\hat{w}_{t+1,0} = {w}_{t+1}$ \;
        \For{$e = 0$ \KwTo $E-1$}{
            $\hat{w}_{t+1,e+1} = \hat{w}_{t+1,e} -  \zeta_t \tilde{\nabla} f_\mathcal A (\hat{w}_{t+1,e})$; // Server uses the auxiliary dataset $\mathcal A$ to correct direction
        }
        $w_{\text{final}} = \hat{w}_{t+1,E}$ and \textbf{Exit the algorithm} \;
    }
}
\end{algorithm}

\section{The \textsc{FedDDPM+} Algorithm}
\label{Sec:5}
In order to tackle the challenge of data heterogeneity in FL, the \textsc{FedDDPM} algorithm is designed to optimize the global model in each round by using the synthesized auxiliary dataset. However, this approach can be computationally expensive for servers with limited computing power, resulting in a delay in broadcasting the next round of the global model to each client for training. This delay can potentially impact the overall efficiency of the FL system. To address these issues, we propose an improved algorithm, \textsc{FedDDPM+}, which optimizes the global model only when slow learning progress is detected.

Specifically, the design of \textsc{FedDDPM+} draws inspiration from two phenomena.
\begin{enumerate}
\item \textbf{Training marginal gains.}  During the initial stages of training, the model can achieve significant performance improvements. However, as the training progresses, it is common for the model to exhibit minimal performance improvements \cite{gulli2017deep, ye2021auxiliary}. This issue can be further exacerbated by the presence of heterogeneous data, resulting in small performance gains at high training costs in later stages. To address this issue, we propose the \textsc{QuickTest} algorithm based on exponential moving averages. This algorithm can identify slow learning progress in the global model and halt FL training early.  By implementing this strategy, we can achieve adequate performance improvements in the global model with less training overhead.
\item \textbf{FL training efficiency.} Although calibrating the global model at each round may be an effective technique to deal with non-IID\cite{karimireddy2020scaffold}, the requirement for additional server training may affect the efficiency of FL. To alleviate this issue, we develop the \textsc{FedDDPM+} algorithm, which detects a learning bottleneck caused by non-IID data, and subsequently uses the auxiliary dataset to optimize the global model for multiple rounds to improve model performance. This approach obviates the requirement for servers to engage in FL training sequentially after model aggregation.
\end{enumerate}
\begin{algorithm}[ht]
\caption{The \textsc{QuickTest} Algorithm.}
\label{alg:algorithm3}
\KwIn{Model $w$; $AvgScore$; $Threshold$; $\gamma$; \textsc{TestSize}}
\KwOut{True; False}
$Score = \textsc{Evaluate}(\textsc{Sampling}(w, \textsc{TestSize}))$ \;

\If{$AvgScore \neq \emptyset$ \textbf{and} $|AvgScore - Score| \leq Threshold$}{
    \Return True \;
}

\If{$AvgScore = \emptyset$}{
    $AvgScore = Score$ \;
}
\Else{
    $AvgScore = Score \cdot \gamma + AvgScore \cdot (1 - \gamma)$ \;
}

\Return False \;
\end{algorithm}

The workflow of \textsc{FedDDPM+} is similar to the \textsc{FedDDPM} discussed in Section \ref{Sec:3}-B, but with a few key differences. Firstly, \textsc{FedDDPM+} does not involve server-side training per round. Secondly, the global model's performance is evaluated every ten rounds by sampling a batch of data and computing their Fréchet Inception Distance (FID) score, which is a measure of the distance between the generated images and the real images. To ensure reliable performance measurements, an exponential moving average-based \textsc{QuickTest} algorithm is used to smooth out fluctuations in the FID score by aggregating past scores with a weighting factor $\gamma$.  If the absolute difference between the current score and the previous averaged score is below a preset threshold, indicating that the model is learning slowly. In such a case, \textsc{FedDDPM+} stops broadcasting the global model, optimizes it using an auxiliary dataset, and terminates training.

In summary, the \textsc{FedDDPM+} algorithm has the benefit of lower training costs compared to \textsc{FedDDPM}. However, its performance may be impacted as the global model is not optimized throughout the entire FL training process.  It is worth mentioning that the convergence analysis of \textsc{FedDDPM+} is a simplified version of \textsc{FedDDPM}, so we do not go into further detail in this regard. In the next section, we conduct experiments to compare the performance of our two proposed algorithms with the current state-of-the-art algorithms.

\section{Experimental}
\label{Sec:6}
\subsection{Experimental Setup}
\subsubsection{\textbf{Model and Datasets}}  
To evaluate the effectiveness of the proposed \textsc{FedDDPM} and \textsc{FedDDPM+} algorithms, we perform image synthesis tasks on three datasets: MNIST\cite{lecun1998mnist}, CIFAR10, and CIFAR100\cite{krizhevsky2009learning}. The MNIST dataset comprises grayscale images of handwritten digits, each with a resolution of 28 × 28 pixels and labeled with one of 10 possible digits. The CIFAR10 dataset contains 60,000 color images with a resolution of 32 × 32 pixels, divided into 10 distinct classes, such as animals (e.g., cats, dogs) and vehicles (e.g., cars, trucks). This dataset introduces more complexity due to the variability in color and object shapes. Finally, the CIFAR100 dataset is significantly more challenging, featuring 100 classes, each containing 600 images. These images are also 32 × 32 pixels in resolution but cover a broader range of categories, including fine-grained distinctions within animal species, types of vehicles, and other objects.

For MNIST, we implement the denoising U-Network-based model $\epsilon_\theta$ includes embedding layers with 500 timestamps. The encoder and decoder each consist of three layers, each containing a residual block. The channel size of each layer is set to 128, 128, and 256, respectively, with the spatial resolution of the feature map on each layer being 28 × 28, 14 × 14, and 7 × 7, respectively. For CIFAR10 and CIFAR100, the denoising model $\epsilon_\theta$ for this dataset includes embedding layers with 1000 timestamps. Each encoder and decoder component of the model consists of four layers, each containing two residual blocks and multi-head self-attention mechanisms. The channel size of each layer is set to 128, 256, 256, and 256, respectively, with the spatial resolution of the feature map on each layer being 32 × 32, 16 × 16, 8 × 8, and 4 × 4, respectively.

To comprehensively evaluate the generalization capabilities of our proposed method, we focus on varying degrees of non-IID partitioning to simulate real-world data distribution scenarios, including shard partitioning and Dirichlet distribution-based partitioning. In the shard partitioning, data samples are sorted by their labels and divided into $2n$ groups, with each client randomly receiving two groups. In the Dirichlet partitioning, we adjust the $\alpha$ values to manage different levels of data heterogeneity, controlling the number and class distribution of data among clients. In our experiments, $\alpha$ is set to 0.1 and 0.3, with lower $\alpha$ values indicating higher heterogeneity.

\subsubsection{\textbf{Metric and (Hyper)parameters}}  
Fréchet Inception Distance \cite{heusel2017gans} as a metric for evaluating the quality of generated images by comparing the distribution of real and generated images. Unlike traditional metrics such as Inception Score\cite{barratt2018note} or pixel-wise metrics like PSNR, FID considers the diversity of features in the generated image. Moreover, FID is a more robust metric and correlates better with human judgment, with smaller values consistently aligning with higher perceived image quality. In our experiments, following the criteria in \cite{ho2020denoising, nichol2021improved}, we computed the FID score using the training set, as is standard practice.

There are several important  hyperparameters in the experiment. The fraction of clients selected in each round is denoted as $p$, taking a value in the range (0, 1]. $T$ represents the number of communication rounds, $K$ refers to the number of local epochs, and $E$ represents the number of epochs performed by the server. $N$ and $B$ denote the total number of clients and batch size, respectively. For the MNIST dataset, we set $T$ = 400, $B$ = 128, $K$ = 3, $E$ = 1 in \textsc{FedDDPM}, and $E=20$ in \textsc{FedDDPM+}, with $p$ = 0.15 and 0.3. For the CIFAR10 and CIFAR100 datasets, we set $T$ = 600, $B$ = 64, $K$ = 5, $E$ = 5 in \textsc{FedDDPM}, and $E=30$ in \textsc{FedDDPM+}, with $p$ = 0.15 and 0.3. The following hyperparameters for test datasets are identical: the number of clients $N$ is set as 100, the learning rate $\eta_t$ is set as $5e^{-5}$, and $\zeta_t$ is set as $2e^{-4}$.

\begin{table*}[htb]
\caption{\textsc{Comparing FID Scores of MNIST Dataset}.}
\label{table:1}
\centering
\resizebox{\linewidth}{!}{ 
\begin{tabular}{|c|c|c|c|c|c|c|}
\hline

\multicolumn{1}{c}{\multirow{3}{*}{Schemes}} & 

\multicolumn{3}{c}{100 clients 15\% participation}                                                               & \multicolumn{3}{c}{100 clients 30\% participation}
\\ \cline{2-7} 
\multicolumn{1}{c}{}                         & \multicolumn{6}{c}{Data Distribution}                       \\ \cline{1-7}                                                                                                                                                           \multicolumn{1}{c}{}                         & \multicolumn{1}{c}{Shard} & \multicolumn{1}{c}{Dirichlet $\alpha = 0.1$} & \multicolumn{1}{c}{Dirichlet $\alpha = 0.3$} & \multicolumn{1}{c}{Shard} & \multicolumn{1}{c}{Dirichlet $\alpha = 0.1$} & \multicolumn{1}{c}{Dirichlet $\alpha = 0.3$} \\
FedAvg                                       & 5.822                     & 4.448                                        & 2.786                                        & 4.868                     & 2.842                                        & 2.377                       \\FedProx & 5.714 &  4.331 & 2.981 & 4.651 & 2.893 & 2.343
\\SCAFFOLD & 4.889 & 3.876 & 2.584 & 4.237 & 2.421 & 2.150\\ 
MoFedSAM & 4.840 & 2.797 & 2.358 & 4.324 & 2.983 & 1.903\\
FedDDPM & \textbf{1.910}& 2.151& 1.997& \textbf{1.842}& \textbf{1.662}& 1.768\\
FedDDPM+& 2.188& \textbf{2.037}& \textbf{1.876}& 1.865& 1.791& \textbf{1.762}      
\end{tabular}
}
\end{table*}

\begin{figure*}[htb]
\centering
\begin{minipage}[t]{0.161\linewidth}
  \centering
  \includegraphics[width=\linewidth]{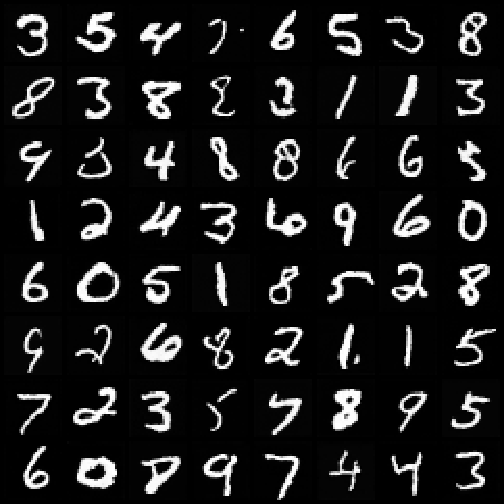}
  \vspace{2pt}
  \centerline{(a) FedAvg}\medskip
\end{minipage}\hfill
\begin{minipage}[t]{0.161\linewidth}
  \centering
  \includegraphics[width=\linewidth]{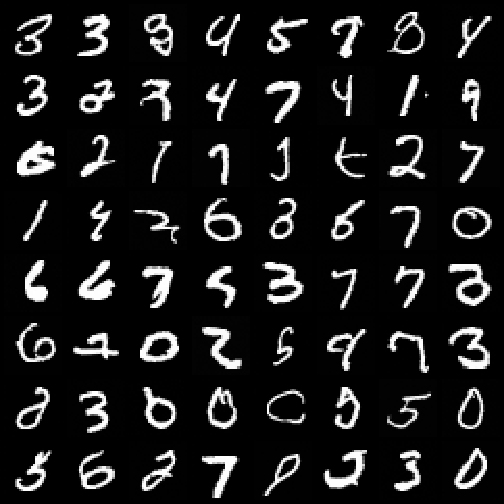}
  \vspace{2pt}
  \centerline{(b) FedProx}\medskip
\end{minipage}\hfill
\begin{minipage}[t]{0.161\linewidth}
  \centering
  \includegraphics[width=\linewidth]{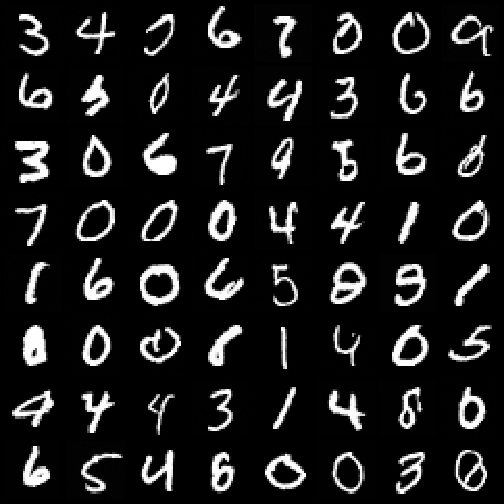}
  \vspace{2pt}
 \centerline{(c) SCAFFOLD}\medskip
\end{minipage}\hfill
\begin{minipage}[t]{0.161\linewidth}
  \centering
  \includegraphics[width=\linewidth]{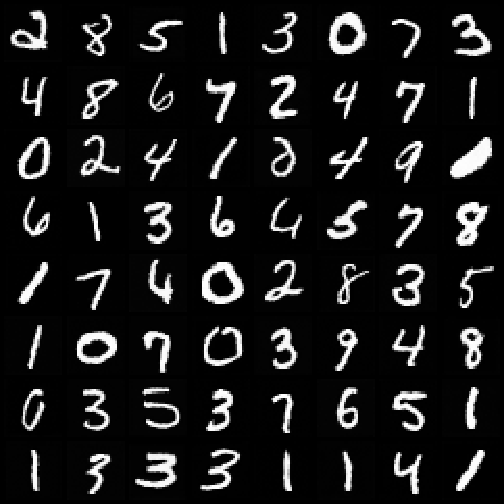}
  \vspace{2pt}
 \centerline{(d) MoFedSAM}\medskip
\end{minipage}\hfill
\begin{minipage}[t]{0.161\linewidth}
  \centering
  \includegraphics[width=\linewidth]{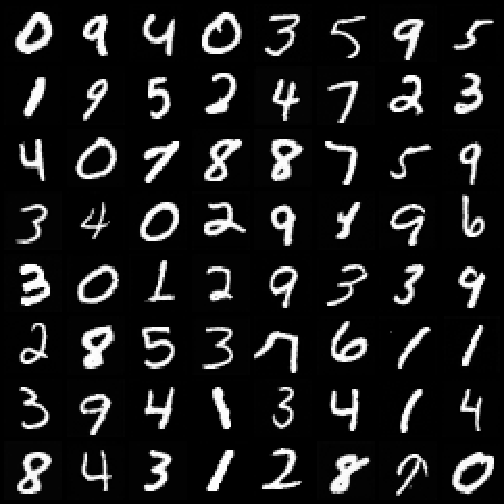}
  \vspace{2pt}
 \centerline{(e) FedDDPM}\medskip
\end{minipage}\hfill
\begin{minipage}[t]{0.161\linewidth}
  \centering
  \includegraphics[width=\linewidth]{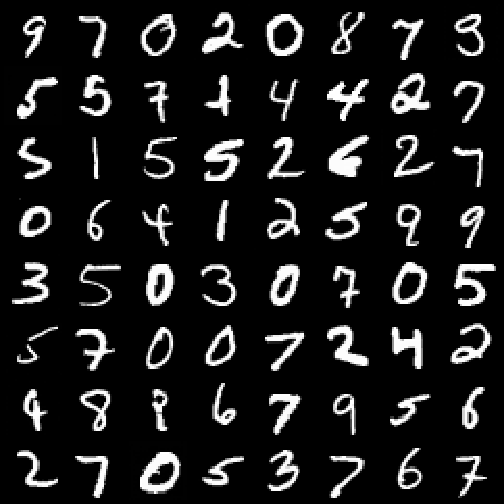}
  \vspace{2pt}
 \centerline{(f) FedDDPM+}\medskip
\end{minipage}
\caption{Synthetic Image Results: Diffusion Models Trained on MNIST with Shard Distribution Using Various FL Algorithms.}
\label{fig:4}
\end{figure*}

In view of the marginal gain of model performance from the size of the auxiliary dataset\cite{pi2023dynafed}, after extensive experimental testing we strategically selected a $|\mathcal A_i| = 0.1 \times m_i $ in each round to balance training efficiency and performance.
In the implementation of \textsc{FedDDPM+}, the \textsc{QuickTest} algorithm uses the hyperparameters $\gamma$ = 0.4, \textsc{Threshold} = 0.2, and \textsc{TestSize} = 500 \cite{hardy2019md}. Unless otherwise explicitly stated, these parameter settings remained unchanged in all subsequent experiments.

\subsubsection{\textbf{Baselines}}  
In addition to the implementation of \textsc{FedDDPM} and its enhanced version, \textsc{FedDDPM+}, we incorporate four baseline methods for comparative analysis. The first baseline, \textsc{FedAvg}, a widely recognized approach, periodically averages updates from individual clients. Additionally, we also include three state-of-the-art algorithms designed to handle non-IID data: \textsc{FedProx}\cite{fedprox}, \textsc{Scaffold}\cite{karimireddy2020scaffold}, and \textsc{MoFedSAM}\cite{qu2022generalized}.
Specifically, \textsc{FedProx} employs local regularization to restrict the magnitude of local model updates. In contrast, \textsc{Scaffold} utilizes personalized local training objectives to correct the direction of local updates. Meanwhile, \textsc{MoFedSAM} combines the SAM optimizer technology with global momentum to ensure the generalization of the global model. These algorithms exhibit notable strengths in addressing non-IID challenges across diverse tasks.

\begin{table*}[h]
\caption{\textsc{Comparing FID Scores of CIFAR10 Dataset}.}
\label{table:2}
\centering
\resizebox{\linewidth}{!}{ 
\begin{tabular}{|c|c|c|c|c|c|c|}
\hline

\multicolumn{1}{c}{\multirow{3}{*}{Schemes}} & 

\multicolumn{3}{c}{100 clients 15\% participation}                                                               & \multicolumn{3}{c}{100 clients 30\% participation}
\\ \cline{2-7} 
\multicolumn{1}{c}{}                         & \multicolumn{6}{c}{Data Distribution}                       \\ \cline{1-7}                                                                                                                                                        \multicolumn{1}{c}{}                         & \multicolumn{1}{c}{Shard} & \multicolumn{1}{c}{Dirichlet $\alpha= 0.1$} & \multicolumn{1}{c}{Dirichlet $\alpha = 0.3$} & \multicolumn{1}{c}{Shard} & \multicolumn{1}{c}{Dirichlet $\alpha = 0.1$} & \multicolumn{1}{c}{Dirichlet $\alpha = 0.3$} \\
FedAvg                              &39.326  & 35.348 & 28.686 & 33.669 & 32.886 & 25.418                  \\FedProx &36.019  & 34.230   & 29.167  & 35.131 & 33.118 & 25.351
\\SCAFFOLD & 29.659  & 33.053 &  28.616 & 27.820 & 30.561 & 24.957\\
MoFedSAM & 35.522 & 31.045 & 29.422 & 28.626 & 30.672 & 23.852\\
FedDDPM & 22.296 & \textbf{21.730} & \textbf{18.089} & \textbf{19.731} & \textbf{18.817} & \textbf{16.354}
\\FedDDPM+ & \textbf{21.508} & 22.512 & 20.531 & 21.468 & 20.753 & 18.741       \end{tabular}
}
\end{table*}

\begin{figure*}[h] 

\begin{minipage}[b]{0.15\linewidth}
  \centering
    \includegraphics[width=3.05cm]{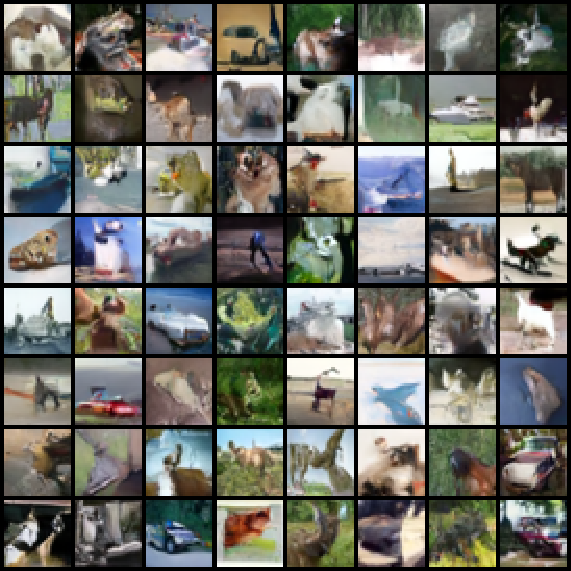}
  \hfill
  \centerline{(a) FedAvg}\medskip
\end{minipage}
\hfill
\begin{minipage}[b]{0.15\linewidth}
  \centering
      \includegraphics[width=3.05cm]{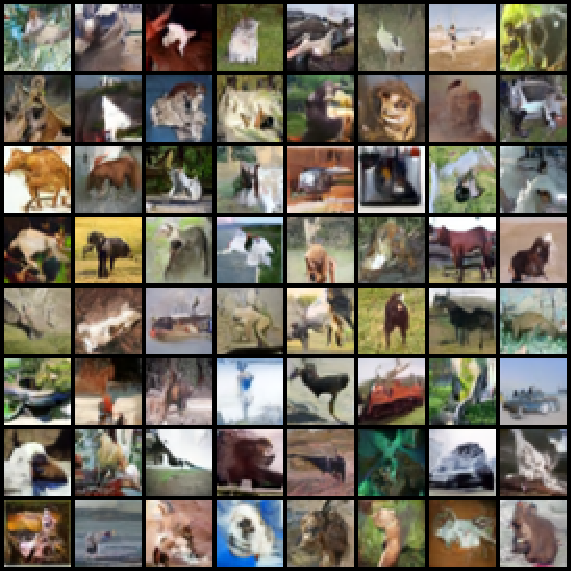}
  \hfill
  \centerline{(b) FedProx}\medskip
\end{minipage}
\hfill
\begin{minipage}[b]{0.15\linewidth}
  \centering
      \includegraphics[width=3.05cm]{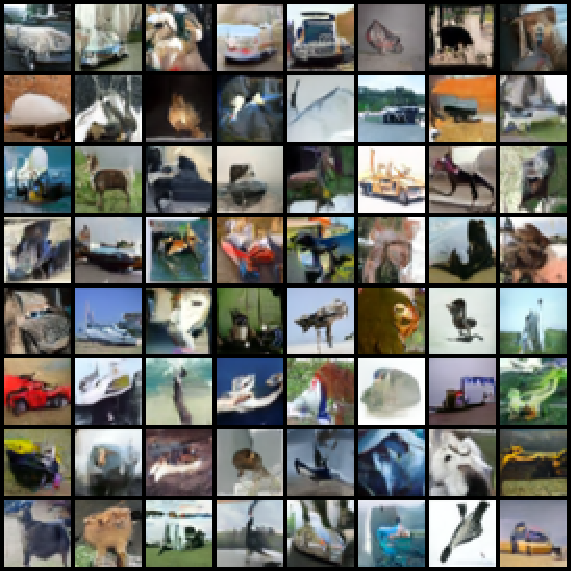}
  \hfill
  \centerline{(c) SCAFFOLD}\medskip
\end{minipage}
\hfill
\begin{minipage}[b]{0.15\linewidth}
  \centering
        \includegraphics[width=3.05cm]{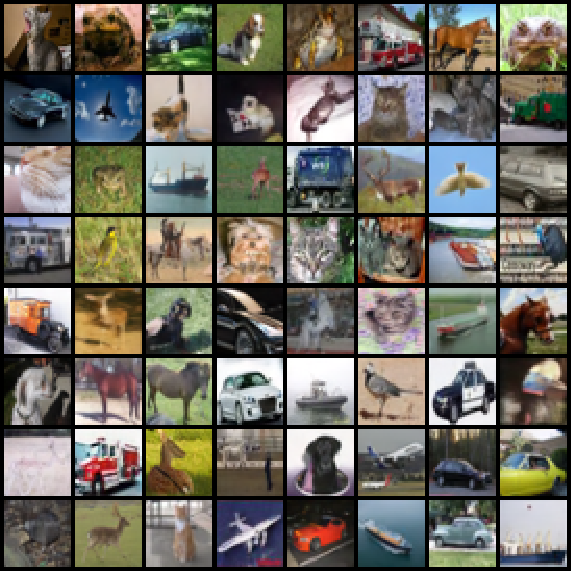}
  \hfill
  \centerline{(d) MoFedSAM}\medskip
\end{minipage}
\hfill
\begin{minipage}[b]{0.15\linewidth}
  \centering
        \includegraphics[width=3.05cm]{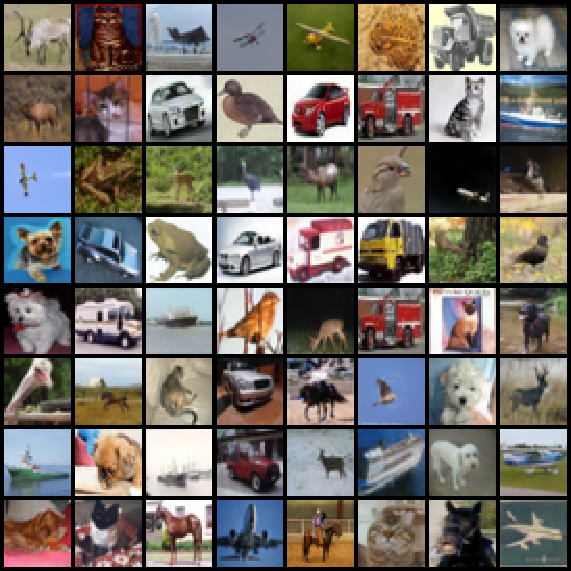}
  \centerline{(e) FedDDPM}\medskip
\end{minipage}
\hfill
\begin{minipage}[b]{0.15\linewidth}
  \centering
        \includegraphics[width=3.05cm]{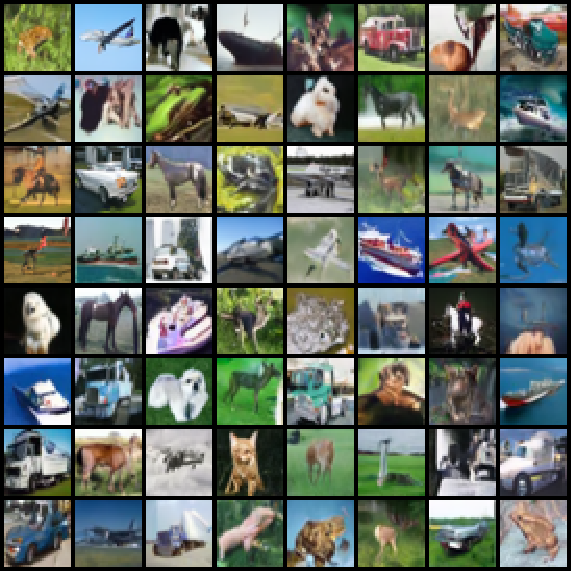}
  \centerline{(f) FedDDPM+}\medskip
\end{minipage}

\caption{Synthetic Image Results: Diffusion Models Trained on CIFAR10 with Dirichlet Distribution  Using Various FL Algorithms.}
\label{fig:5}
\end{figure*}
\subsection{Results on MNIST}
In the \textbf{\textsc{Warmup}} stage of MNIST, every client is instructed to train the model for 200 epochs\footnote{The number of epochs depends on the dataset's complexity. For MNIST, 200 epochs suffice. For more complex datasets, clients may need more epochs to achieve a good fit.} to accurately fit its local data distribution. We conduct experiments under three levels of non-IID distributions: Shard and Dirichlet with $\alpha$ = 0.1 and $\alpha$ = 0.3, with client ratios of 0.15 and 0.3. For the \textsc{FedDDPM+} algorithm, it meets the \textsc{QuickTest} criterion and undergoes fine-tuning after training halts at rounds 160, 150, 170, 130, 130, and 120 for each configuration, significantly reducing training overhead.

Table \ref{table:1} reports the FID scores of the MNIST dataset for \textsc{FedAvg}, \textsc{FedProx}, \textsc{SCAFFOLD}, \textsc{FedSAM},
\textsc{FedDDPM} and \textsc{FedDDPM+}. It can be seen that our proposed algorithm \textsc{FedDDPM} and \textsc{FedDDPM+} outperform other baseline algorithms significantly under different degrees of non-IID data. Specifically, we observe that under the Shard data distribution with the client ratio of 0.15, \textsc{FedDDPM} achieves the best FID value of 1.910, whereas the best-performing baseline algorithm, SCAFFOLD, only achieves an FID value of 4.889. These results demonstrate the superiority of using the auxiliary dataset for model calibration. In addition, \textsc{FedDDPM+} also achieves a good FID value of 2.188 with a much lower training overhead, further validating the effectiveness of our proposed algorithm. A similar phenomenon can be observed in other configurations, and our proposed algorithm consistently outperforms other baselines. Figure \ref{fig:4} depicts the images generated under different FL schemes when the client ratio is 0.15 under the shard distribution, and we can see that our proposed method generates better quality images. The experimental results demonstrate that our proposed algorithms can significantly mitigate the impact of non-IID data on the performance of FL.

\subsection{Result on CIFAR10}
We perform the experiment on CIFAR10. Prior to FL training, all clients undergo a pre-training phase of 400 epochs in \textbf{\textsc{Warmup}} to ensure that their models adequately capture the local distribution. We explore three levels of non-IID distributions, including Shard and Dirichlet distributions with $\alpha$ values of 0.1 and 0.3, along with client ratios of 0.15 and 0.3. In each training configuration, the \textsc{FedDDPM+} algorithm successfully meets the termination condition of the \textsc{QuickTest} and concludes training at rounds 280, 310, 320, 250, 270, and 280, respectively. Compared to other benchmarks, \textsc{FedDDPM+} ensures competitive training costs.

Table \ref{table:2} reports the FID scores of all configurations of the CIFAR10 dataset. We observed that in all configurations, \textsc{FedDDPM} and \textsc{FedDDPM+} significantly outperform the other schemes. Furthermore, Our experiments indicate that a decrease in the Dirichlet parameter $\alpha$ is accompanied by a corresponding drop in the performance of all algorithms. Nevertheless, our proposed \textsc{FedDDPM} and \textsc{FedDDPM+} algorithms are still significantly better than others, confirming that optimizing model learning via auxiliary datasets is indeed an effective approach. We also observe that \textsc{FedDDPM}+ achieves slightly lower performance than \textsc{FedDDPM} but with less training overhead. Figure \ref{fig:5} depicts the images generated by various FL schemes with a client ratio of 0.15 under the shard distribution. The results demonstrate that \textsc{FedDDPM} and \textsc{FedDDPM+} exhibit superior image quality compared to other methods on CIFAR10.

\begin{table*}[h]
\caption{\textsc{Comparing FID Scores of CIFAR100 Dataset}.}
\label{table:3}
\centering
\resizebox{\linewidth}{!}{ 
\begin{tabular}{|c|c|c|c|c|c|c|}
\hline

\multicolumn{1}{c}{\multirow{3}{*}{Schemes}} & 

\multicolumn{3}{c}{100 clients 15\% participation}                                                               & \multicolumn{3}{c}{100 clients 30\% participation}
\\ \cline{2-7} 
\multicolumn{1}{c}{}                         & \multicolumn{6}{c}{Data Distribution}                       \\ \cline{1-7}                                                                                                                                                        \multicolumn{1}{c}{}                         & \multicolumn{1}{c}{Shard} & \multicolumn{1}{c}{Dirichlet $\alpha= 0.1$} & \multicolumn{1}{c}{Dirichlet $\alpha = 0.3$} & \multicolumn{1}{c}{Shard} & \multicolumn{1}{c}{Dirichlet $\alpha = 0.1$} & \multicolumn{1}{c}{Dirichlet $\alpha = 0.3$} \\
FedAvg                              &37.711  & 30.281 & 30.674 & 37.829 & 31.105 & 28.667                 \\FedProx &40.075  & 32.732   & 32.878  & 38.426 & 34.414 & 32.580
\\SCAFFOLD & 38.415  & 31.862 &  31.456 & 37.154& 29.670 & 30.161
\\
MoFedSAM & 37.646 & 30.442 & 28.719 & 36.817 & 28.502 & 28.923
\\FedDDPM & \textbf{21.818} & 23.416 & \textbf{17.139} & \textbf{19.096} & \textbf{24.026} & \textbf{20.196}
\\FedDDPM+ & 23.612 & \textbf{22.007} & 20.793 & 22.996 & 27.643 & 24.554       \end{tabular}
}
\end{table*}
\begin{figure*}[h] 

\begin{minipage}[b]{0.15\linewidth}
  \centering
  \centerline{\includegraphics[width=3.05cm]{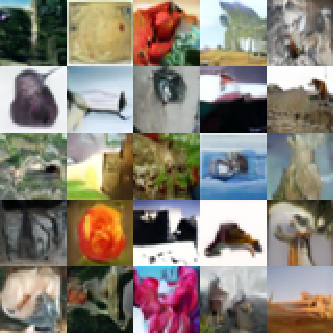}}
  \centerline{(a) FedAvg}\medskip
\end{minipage}
\hfill
\begin{minipage}[b]{0.15\linewidth}
  \centering
  \centerline{\includegraphics[width=3.05cm]{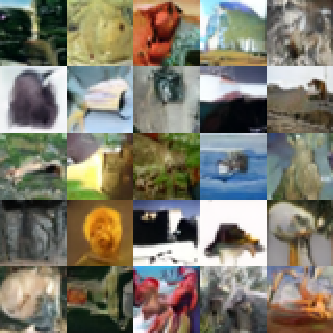}}
  \centerline{(b) FedProx}\medskip
\end{minipage}
\hfill
\begin{minipage}[b]{0.15\linewidth}
  \centering
  \centerline{\includegraphics[width=3.05cm]{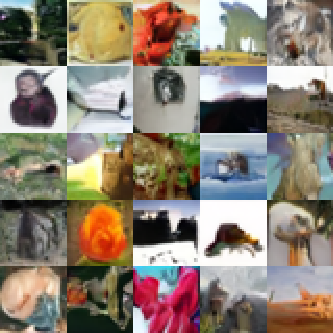}}
  \centerline{(c) SCAFFOLD}\medskip
\end{minipage}
\hfill
\begin{minipage}[b]{0.15\linewidth}
  \centering
  \centerline{\includegraphics[width=3.05cm]{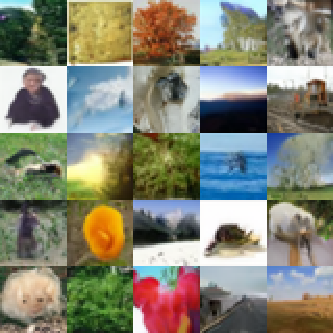}}
  \centerline{(d) MoFedSAM}\medskip
\end{minipage}
\hfill
\begin{minipage}[b]{0.15\linewidth}
  \centering
  \centerline{\includegraphics[width=3.05cm]{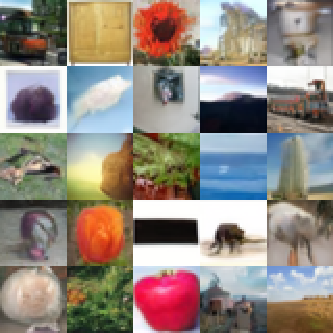}}
  \centerline{(e) FedDDPM}\medskip
\end{minipage}
\hfill
\begin{minipage}[b]{0.15\linewidth}
  \centering
  \centerline{\includegraphics[width=3.05cm]{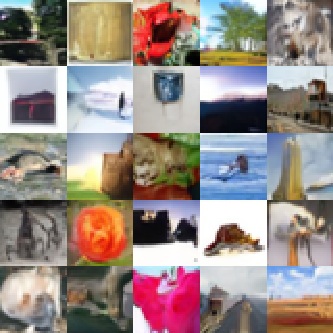}}
  \centerline{(e) FedDDPM+}\medskip
\end{minipage}
\caption{Synthetic Image Results: Diffusion Models Trained on CIFAR100 with Dirichlet Distribution Using Various FL Algorithms.}
\label{fig:6}
\end{figure*}

\subsection{Result on CIFAR100}
We repeat the experiments on the CIFAR100 dataset, which is a more challenging task. Preceding the training stage, every client undergoes a 600-epoch pre-training process to construct the auxiliary dataset. Under the same non-IID configuration as mentioned earlier, \textsc{FedDDPM+} satisfies the conditions of the \textsc{QuickTest} algorithm at rounds 310, 330, 280, 290, 300, and 270, respectively.

The experimental results in Table \ref{table:3} demonstrate that, even on the more challenging CIFAR100 data, both \textsc{FedDDPM} and \textsc{FedDDPM+} consistently outperform other benchmark algorithms under different data distributions and client participation rates. This underscores the robust generalization performance and practicality of our proposed method. Figure \ref{fig:6} shows images generated by various algorithms using the same seed under a Dirichlet distribution with $\alpha=0.3$ and a client ratio of 30\%. Similarly, it is evident that \textsc{FedDDPM} excels in producing high-quality images.

\subsection{Further Experiments and Analysis}
\begin{table}[h]
\centering
\setlength{\tabcolsep}{4pt}
\begin{tabular}{lccc}
\hline
\textbf{} & \textbf{MNIST Shard} & \textbf{CIFAR10} & \textbf{CIFAR100} \\
 &  & \textbf{Dir. $\alpha=0.1$} & \textbf{Dir. $\alpha=0.3$} \\
\hline
\multicolumn{4}{c}{\textbf{FID of synthetic auxiliary dataset (for reference)}} \\
Synthetic Aux. Dataset & 1.344  & 17.435 & 14.420 \\
\hline
Training w/ Aux. Data & 4.204 & 47.332 & 49.434 \\
FedAvg  & 5.822  &  35.348 & 30.674 \\
FedDDPM  & \textbf{1.910}  & \textbf{21.730} & \textbf{17.139} \\
\hline
\end{tabular}
\caption{\textsc{FID Scores: Centralized Training with Synthetic Auxiliary Datasets vs. FL Training}.}
\label{tab:non_iid_performance}
\end{table}

\begin{figure}[htbp]
    \centering
    \begin{minipage}[b]{0.49\linewidth}
    \centering
    \includegraphics[width=4.4cm]{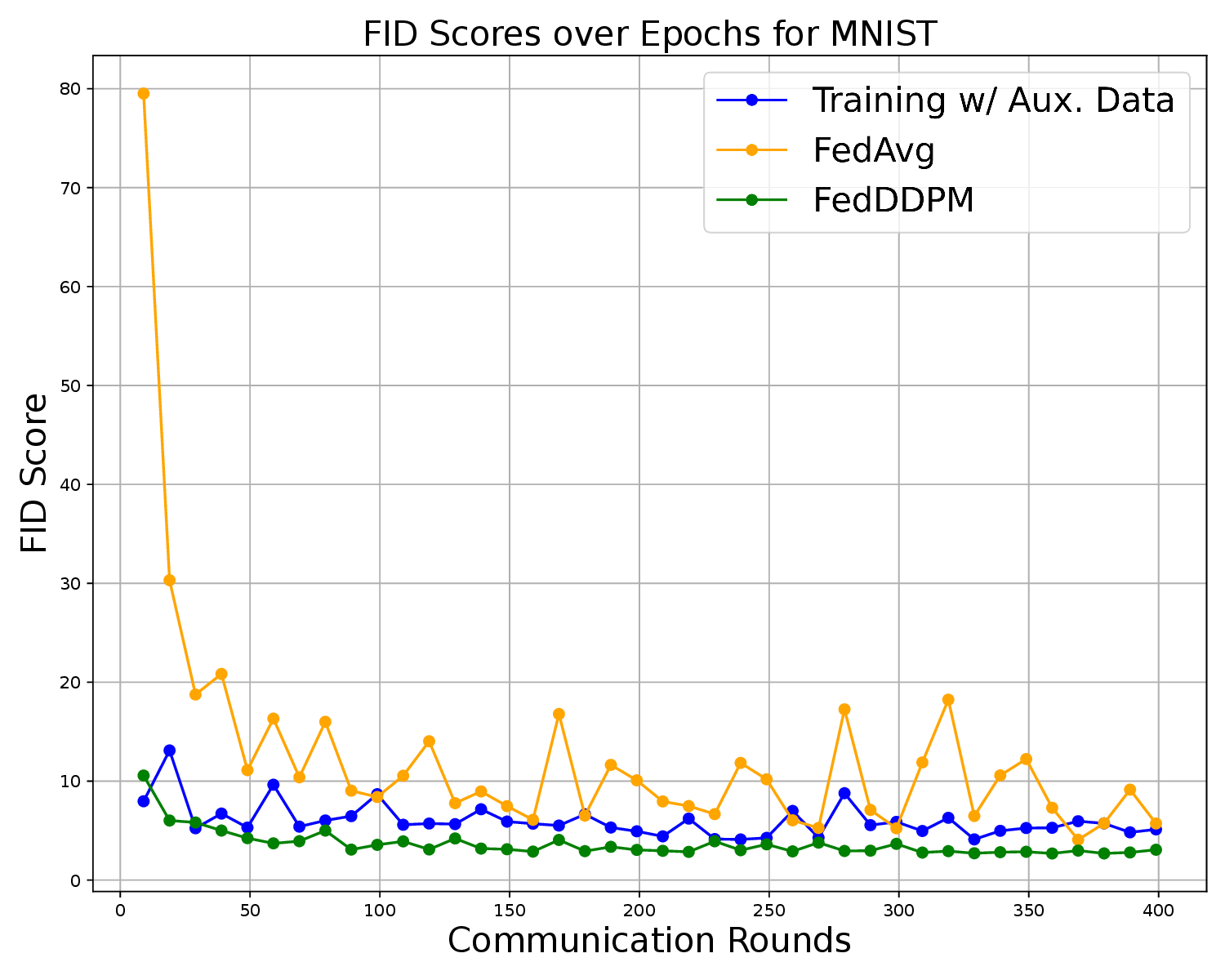}
    \label{fig:mnist_fid}
    \end{minipage}
    \begin{minipage}[b]{0.49\linewidth}
        \centering
        \includegraphics[width=4.4cm]{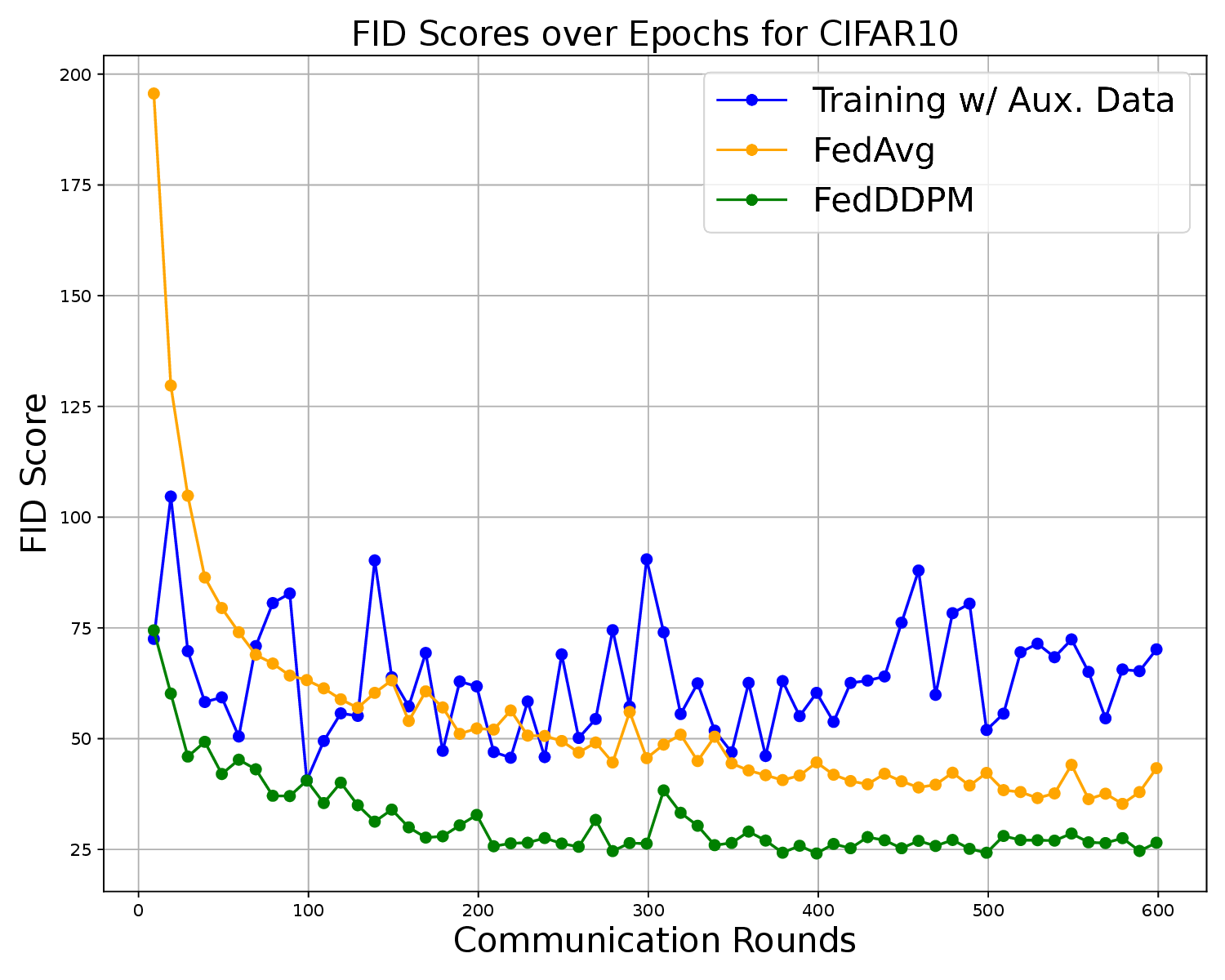}
        \label{fig:cifar_fid}
    \end{minipage}
    \hfill 
    \caption{FID scores across epochs for the MNIST and CIFAR10 datasets.}
    \label{fig:fid_over_epoch}
\end{figure}
\subsubsection{\textbf{Assessing the Necessity of FL Training}}
To evaluate the necessity of FL aggregation, we conduct further analyses in a scenario where only synthetic auxiliary datasets are utilized for the centralized training of diffusion models. This investigation aims to determine whether centralized training exclusively dependent on auxiliary datasets can yield proficient models. For privacy and comparability, we generate auxiliary datasets that match the size of the original datasets, maintaining previous configurations at different levels of non-IID across all three datasets.

As shown in Table \ref{tab:non_iid_performance}, the experimental results indicate that, even with 15\% client participation, both \textsc{FedAvg} and \textsc{FedDDPM} outperform centralized synthetic dataset training in the vast majority of cases, particularly in more complex tasks. Furthermore, as illustrated in Figure \ref{fig:fid_over_epoch}, we track the trend of FID changes every 10 rounds, revealing that \textsc{FedDDPM} exhibits a more rapid decline in FID. The inferior performance of centralized training using auxiliary datasets arises from the performance degradation observed during secondary training on these datasets, similar to the degradation noted when training on the original dataset. While auxiliary datasets are invaluable for simulating real-world data conditions, they cannot fully substitute real datasets in centralized model training.
\subsubsection{\textbf{Comparing Server-Side vs. Client-Side Auxiliary Data Usage}}
\begin{figure}[h]
\centering
\includegraphics[width=80mm]{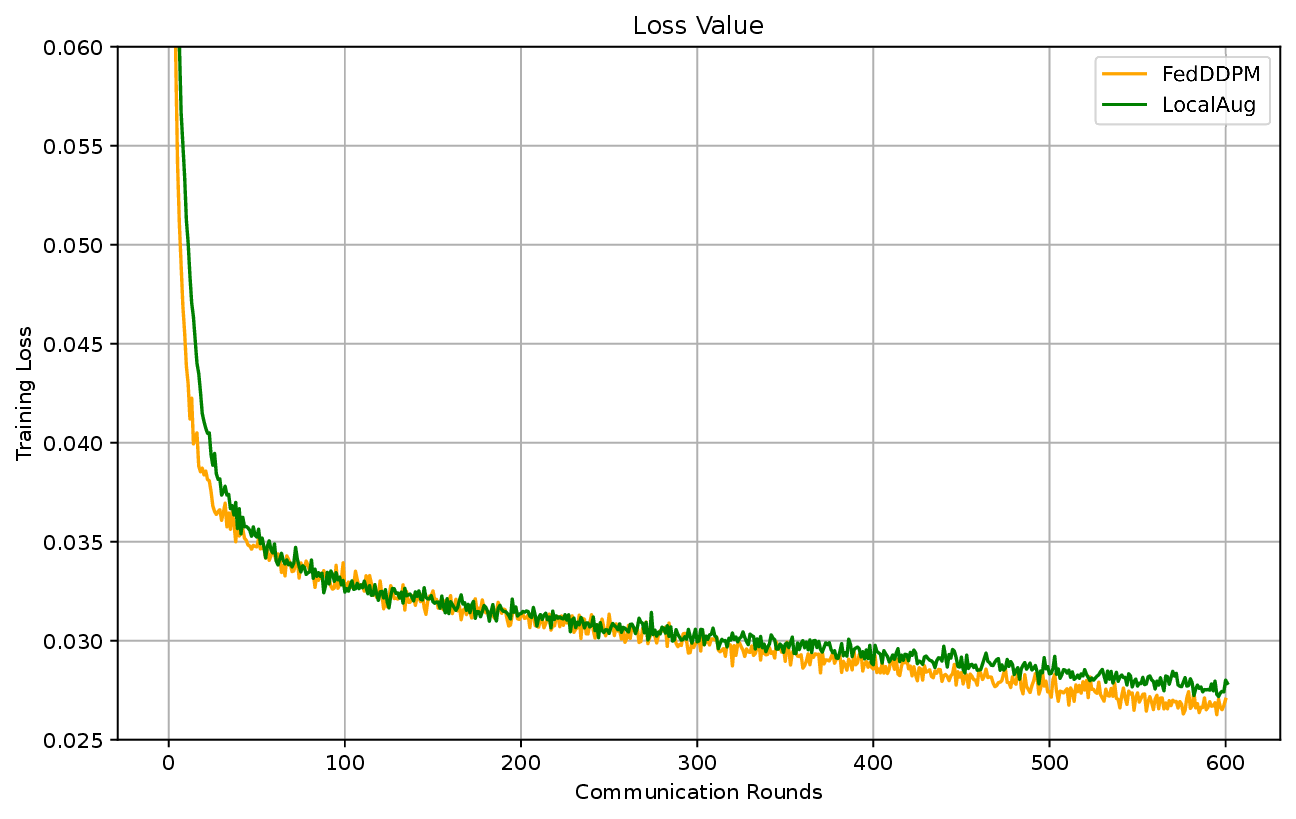}
\caption{Model loss comparison: Using auxiliary data on the client vs. on the server, \textsc{FedDDPM} shows a faster loss reduction trend.}
\label{fig7}
\end{figure}

In this experiment, we evaluate the effectiveness of using auxiliary datasets to optimize FL training on the client and server sides, respectively. We select 20 clients and provide each of them with 500 CIFAR100 non-IID images generated from a Dirichlet distribution, where the concentration parameter $\alpha$ is set to 0.1. Besides our proposed \textsc{FedDDPM} method, we evaluate another method named \textsc{LocalAug}, which distributes the auxiliary dataset evenly to the local systems to correct biased local updates, as in \cite{hardy2019md,zhang2022fine}. However, since the clients have access to simulated global data distributions, this method poses substantial privacy concerns as mentioned earlier. In contrast, our \textsc{FedDDPM} algorithm uses auxiliary data solely for model calibration on the trusted server. To ensure a fair assessment, we employ the same synthetic auxiliary dataset for both methods, comprising 2000 samples. Additionally, we include \textsc{FedAvg} and the best-performing \textsc{MoFedSAM} as FL baselines. We monitor the training loss versus the number of communication rounds, as shown in Figure \ref{fig7}. The experimental results demonstrate that \textsc{FedDDPM} achieves faster initial loss reduction, consistent with our theoretical analysis.

\subsubsection{\textbf{Privacy and Computational Cost Analysis}}
Unlike methods that distribute data to clients, which increase privacy risks \cite{yoon2020fedmix}, \textsc{FedDDPM} introduces a trusted central server to safeguard data privacy. \textsc{FedDDPM} uses synthetic datasets to correct the model on the server side. However, this approach adds computational overhead. To address this, \textsc{FedDDPM}+ mitigates the impact through a dynamic evaluation mechanism that fine-tunes the model when training slows down. For example, for CIFAR100, \textsc{FedDDPM}+ converges in around 300 global rounds, reducing the number of global rounds by half compared to \textsc{FedAvg}. Although it requires additional local training during a warmup phase, this can be done in parallel without aggregation delays. As a result, \textsc{FedDDPM}+ maintains competitive efficiency compared to other algorithms.

\section{Conclusion}
\label{Sec:7}
In summary, our paper introduces \textsc{FedDDPM}, a novel learning method tailored to enhance the generative capabilities of diffusion models within distributed heterogeneous data environments. By leveraging a synthetic auxiliary dataset, our approach calibrates the model after each round of aggregation, effectively counteracting the detrimental effects of data heterogeneity on model performance. Additionally, we present \textsc{FedDDPM+}, an enhanced algorithm that identifies slow learning progress and utilizes the auxiliary dataset for one-shot model optimization, thus reducing additional training overhead. Experimental results underscore the superior performance of both \textsc{FedDDPM} and \textsc{FedDDPM+} when compared to state-of-the-art FL algorithms across varying degrees of non-IID data distribution. 

Currently, applications of FL combined with diffusion models show great promise, particularly in generating synthetic medical data. As diffusion models in federated learning are still developing, future work can explore integrating differential privacy mechanisms to further enhance their potential for securely handling sensitive tasks.

\begin{appendices}
\section{Proofs Of Lemmas \ref{lem:1}, \ref{lem:2}}       
\label{section:appendixA}
\subsection{Proof of Lemma \ref{lem:1}}
\begin{proof}
Due to the L-smooth assumption in Assumption \ref{assumption:1}, we have
\begin{align}
\label{aeq:1}
\mathbb{E}f(\hat{w}_{t+1}) = & \mathbb{E}f\left({w}_{t}-\frac{1}{n} \sum_{i\in Z_t} \sum_{\tau=0}^{K-1} \zeta_t \Tilde{\nabla}f_i(w_{t,\tau}^i)\right)\notag\\ 
\leq & \mathbb{E}f(w_t) + \mathbb{E}\underbrace{\langle \nabla f(w_t), -\frac{1}{n} \sum_{i\in Z_t} \sum_{\tau=0}^{K-1} \zeta_t \Tilde{\nabla}f_i(w_{t,\tau}^i)\rangle}_{A_1} \notag\\&+ \underbrace{\frac{L}{2}\mathbb{E} \left\| \frac{1}{n} \sum_{i \in Z_t} \sum_{\tau=0}^{K-1}\zeta_t \Tilde{\nabla}f_i(w_{t,\tau}^i)\right\|^2.}_{A_2}
\end{align}
Next, we study the upper bounds of $A_1$ and $A_2$, respectively.
\begin{align}
\label{aeq:2}
A_1 &= \mathbb{E}\langle \nabla f (w_t), -\frac{1}{n} \sum_{i\in Z_t} \sum_{\tau=0}^{K-1}\zeta_t\Tilde{\nabla}f_i(w_{t,\tau}^i)\rangle \notag\\ = &\mathbb{E}\langle\nabla f(w_t), -\frac{1}{n}\sum_{i \in [N]} \mathbf{1}(i \in Z_t) \sum_{\tau=0}^{K-1}\zeta_t\Tilde{\nabla} f_i(w_{t,\tau}^i)\rangle \notag \\ \overset{\text{(a)}}{=} &\mathbb{E}\langle\nabla f(w_t), -\frac{1}{N}\sum_{i \in [N]}\sum_{\tau=0}^{K-1}\zeta_t\nabla f_i(w_{t,\tau}^i)\rangle \notag \\ =&\mathbb{E}\langle \nabla f(w_t), -\frac{1}{N} \sum_{i \in [N]} \sum_{\tau=0}^{K-1}\zeta_t \nabla f_i(w_{t,\tau}^i) + \zeta_t K \nabla f(w_t) \notag\\ &- \zeta_t K \nabla f(w_t)\rangle
\notag\\=&\mathbb{E} \langle \nabla f(w_t), -\frac{\zeta_t}{N} \sum_{i \in [N]} \sum_{\tau=0}^{K-1}(\nabla f_i(w_{t,\tau}^i) - \nabla f_i(w_t))\rangle\notag\\&-\zeta_t K\mathbb{E}\left\|\nabla f(w_t) \right\|^2 \notag\\=&
-\zeta_t K \mathbb{E}\left\|\nabla f(w_t) \right\|^2 + \mathbb{E}\langle \sqrt{\zeta_t K} \nabla f(w_t), \notag\\
&-\frac{\sqrt{\zeta_t}}{N\sqrt{K}} \sum_{i \in [N]} \sum_{\tau=0}^{K-1}(\nabla f_i(w_{t,\tau}^i)-\nabla f_i(w_t))\rangle
\notag\\ \overset{\text{(b)}}{=}& -\frac{\zeta_t K}{2}\mathbb{E}\left\| \nabla f(w_t)\right\|^2 -
\frac{\zeta_t}{2 N^2 K} \mathbb{E}\left\|\sum_{i \in [N]} \sum_{\tau=0}^{K-1}\nabla f_i(w_{t,\tau}^i) \right\|^2 \notag\\&
+ \frac{\zeta_t}{2 N^2 K} \mathbb{E}\left\|\sum_{i \in [N]} \sum_{\tau=0}^{K-1}(\nabla f_i(w_{t,\tau}^i)-\nabla f_i(w_t))\right\|^2
\notag \\ \overset{\text{(c)}}{\leq} & -\frac{\zeta_t K}{2} \mathbb{E}\left\|\nabla f(w_t) \right\|^2-\frac{\zeta_t}{2 N^2K}\mathbb{E}\left\| \sum_{i \in [N]} \sum_{\tau=0}^{K-1} \nabla f_i(w_{t,\tau}^i)\right\|^2 \notag \\ &+ \frac{\zeta_t}{2N}\sum_{i \in [N]} \sum_{\tau=0}^{K-1}\mathbb{E}\left\|\nabla f_i(w_{t,\tau}^i)-\nabla f_i(w_t) \right\|^2 \notag \\ \overset{(d)}{\leq}& -\frac{\zeta_t K}{2} \mathbb{E}\left\|\nabla f(w_t) \right\|^2-\frac{\zeta_t}{2N^2K}\mathbb{E}\left\| \sum_{i \in [N]} \sum_{\tau=0}^{K-1} \nabla f_i(w_{t,\tau}^i)\right\|^2 \notag \\ &+ \frac{\zeta_t L^2}{2N}\sum_{i \in [N]} \sum_{\tau=0}^{K-1}\mathbb{E}\underbrace{\left\|w_{t,\tau}^i-w_t \right\|^2,}_{A_3}
\end{align}

where the equality in $(a)$ is established by Assumption \ref{assumption:2} and $\mathbb{E}[\mathbf{1}(i \in Z_t)] = \frac{n}{N}
$, while in $(b)$ it arises from the rearrangement of the inner product $\langle a, b \rangle = \frac{1}{2}\|a\|^2 + \frac{1}{2}\|b\|^2 - \frac{1}{2}
\|a-b\|^2$ for any pair of vectors $a$ and $b$. The validity of $(c)$ is rooted in the Cauchy-Schwarz inequality, and $(d)$ holds as a consequence of Assumption \ref{assumption:1}.
\begin{align}
\label{aeq:3}
 A_3 &= \mathbb{E}\left\|w_{t,\tau}^i-w_t\right\|^2 \notag\\=&\mathbb{E}
 \left\|w_{t,\tau-1}^i-w_t -\zeta_t\Tilde{\nabla} f_i(w_{t,\tau-1}^i)\right\|^2\notag\\
=&\mathbb{E}\left\|w_{t,\tau-1}^i-w_t-\zeta_t(\Tilde{\nabla} f_i(w_{t,\tau-1}^i)-\nabla f_i(w_{t,\tau-1}^i))\notag\right.\\&\left. + \zeta_t(\nabla f_i(w_{t,\tau-1}^i)
 -\nabla f_i(w_t) + \nabla f_i(w_t) \notag\right.\\&\left.  - \nabla f(w_t) + \nabla f(w_t))\right\|^2\notag\\ \overset{\text{(a)}}{=}& \mathbb{E}\left\|w_{t,\tau-1}^i-w_t-\zeta_t(\nabla f_i(w_{t,\tau-1}^i)-\nabla f_i(w_t) +\nabla f_i(w_t)\notag\right.\\&\left.-\nabla f(w_t)+\nabla f(w_t))\right\|^2\notag\\&+\mathbb{E}\left\|\zeta_t(\Tilde{\nabla}f_i(w_{t,\tau-1}^i)-\nabla f_i(w_{t,\tau-1}^i))\right\|^2\notag\\\overset{\text{(b)}}{=}&(1+\frac{1}{2K-1})\mathbb{E}\left\|w_{t,\tau-1}^i-w_{t}\right\|^2+2K\zeta_t^2\mathbb{E}\|\nabla f_i(w_{t,\tau-1}^i)\notag\\&-\nabla f_i(w_t)+\nabla f_i(w_t) -\nabla f(w_t)+\nabla f(w_t)\|^2\notag\\&+\mathbb{E}\|\zeta_t(\Tilde{\nabla} f_i(w_{t,\tau-1}^i)-\nabla f_i(w_{t,\tau-1}^i))\|^2\notag\\\overset{\text{(c)}}{=}&(1+\frac{1}{2K-1})\mathbb{E}\|w_{t,\tau-1}^i-w_t\|^2
 + \zeta_t^2 \sigma_l^2+6K\zeta_t^2 \sigma_g^2  \notag\\&+ 6K \zeta_t^2 \|\nabla f(w_t)\|^2+6K\zeta_t^2\mathbb{E}\|\nabla f_i(w_{t,\tau-1}^i)-\nabla f_i(w_t)\|^2\notag\\ \overset{\text{(d)}}{=}&(1+\frac{1}{2K-1}+6K\zeta_t^2L^2)\mathbb{E}\|w_{t,\tau-1}^i-w_t\|^2+\zeta_t^2\sigma_l^2+\notag\\&6K\zeta_t^2\sigma_g^2 +6K\zeta_t^2\|\nabla f(w_t)\|^2
 \notag\\ \overset{\text{(e)}}\leq&(1+\frac{1}{K-1})\mathbb{E}\|w_{t,\tau-1}^i-w_t\|^2+\zeta_t^2\sigma_l^2+6K\zeta_t^2\sigma_g^2\notag\\&+6K\zeta_t^2\|\nabla f(w_t)\|^2.
 \end{align}
where Assumption \ref{assumption:2} ensures equality in $(a)$. By applying the inequality $\|a+b\|^2 \leq (1+\frac{1}{k})\|a\|^2+(k+1)\|b\|^2$ to all vectors $a$ and $b$, we establish $(b)$, where $k=2K-1$. Equations $(c)$ are supported by the Cauchy-Schwarz inequality and Assumption \ref{assumption:3}, while equation $(d)$ holds as a result of Assumption \ref{assumption:1}. Under the constraints specified in Theorem \ref{theorem:1}, it is evident that inequality $(e)$ is satisfied.

Let $\mathbb{E}\|w_{t,\tau}^i -w_t\|^2=\delta_\tau$, we obtain the following recursive inequality 
\begin{align}
\label{aeq:4}
\delta_\tau \leq& (1+ \frac{1}{K-1}) \delta_{\tau-1}+\zeta_t^2\sigma_l^2+6K\zeta_t^2\sigma_g^2+6K\zeta_t^2\mathbb{E}\left\|\nabla f(w_t)\right\|^2 \notag\\ \leq& (1+\frac{1}{K-1})\big[(1+ \frac{1}{K-1}) \delta_{\tau-2}+\zeta_t^2\sigma_l^2+6K\zeta_t^2\sigma_g^2\notag\\&+6K\zeta_t^2\mathbb{E}\left\|\nabla f(w_t)\right\|^2\big] + \zeta_t^2\sigma_l^2+6K\zeta_t^2\sigma_g^2\notag\\&+6K\zeta_t^2\mathbb{E}\left\|\nabla f(w_t)\right\|^2 \notag \\ \leq&(1+\frac{1}{K-1})^\tau\delta_0+\sum_{j=0}^{\tau-1}(1+\frac{1}{K-1})^j[\zeta_t^2\sigma_l^2+6K\zeta_t^2\sigma_g^2\notag\\&+6K\zeta_t^2\mathbb{E}\left\|\nabla f(w_t)\right\|^2]
\notag\\ =& [(1+\frac{1}{K-1})^\tau-1] (K-1)[\zeta_t^2\sigma_l^2+6K\zeta_t^2\sigma_g^2\notag\\&+6K\zeta_t^2\mathbb{E}\left\|\nabla f(w_t)\right\|^2]
\notag\\ \leq& [(1+\frac{1}{K-1})^K-1] (K-1)[\zeta_t^2\sigma_l^2+6K\zeta_t^2\sigma_g^2\notag\\&+6K\zeta_t^2\mathbb{E}\left\|\nabla f(w_t)\right\|^2] \notag\\ \leq& 3K \zeta_t^2 (\sigma_l^2+6K\sigma_g^2) + 18 K^2 \zeta_t^2 \mathbb{E}\left\|\nabla f(w_t)\right\|^2,
\end{align}
where the last inequality uses the fact that $(1+\frac{1}{K-1})^K \leq 4$ for $K >$  1.
Substitute (\ref{aeq:4}) into  (\ref{aeq:2}), we have
\begin{align}
\label{aeq:5}
A_1& \leq -\frac{\zeta_t K}{2}\mathbb{E}\left\|\nabla f(w_t)\right\|^2- \frac{\zeta_t}{2N^2K}\mathbb{E}\left\|\sum_{i \in [N]} \sum_{\tau=0}^{K-1} \nabla f_i(w_{t,\tau}^i)\right\|^2  \notag \\ &+ \frac{\zeta_t L^2}{2N} \sum_{i \in [N]} \sum_{\tau=0}^{K-1}(3K\zeta_t^2(\sigma_l^2+6K\sigma_g^2)  \notag \\ &+18K^2\zeta_t^2\mathbb{E}\left\|\nabla f(w_t)\right\|^2)
\notag\\ \leq & -K\zeta_t(\frac{1}{2}-9K^2\zeta_t^2L^2) \mathbb{E}\left\|\nabla f(w_t)\right\|^2 \notag \\ &- \frac{\zeta_t}{2N^2K}\mathbb{E}\|\sum_{i \in [N]} \sum_{\tau=0}^{K-1} \nabla f_i(w_{t,\tau}^i)\|^2 + \frac{3}{2} K^2 \zeta_t^3 L^2 (\sigma_l^2+6K\sigma_g^2).
\end{align}
Now, we continue to bound $A_2$,
\begin{align}
\label{aeq:6}
A_2 =& \frac{L}{2} \mathbb{E}\left\| \frac{1}{n} \sum_{i \in Z_t} \sum_{\tau=0}^{K-1}\zeta_t \Tilde{\nabla}f_i(w_{t,\tau}^i)\right\|^2
\notag\\=& \frac{\zeta_t^2 L}{2n^2}\mathbb{E}\left\|\sum_{i \in Z_t}\sum_{\tau=0}^{K-1}(\Tilde{\nabla} f_i(w_{t,\tau}^i)-\nabla f_i(w_{t,\tau}^i)+\nabla f_i(w_{t,\tau}^i))\right\|^2
\notag\\ \overset{(a)}{=}& \frac{\zeta_t^2 L}{2n^2}\mathbb{E}\left\|\sum_{i \in Z_t} \sum_{\tau=0}^{K-1}(\Tilde{\nabla}f_i(w_{t,\tau}^i)-\nabla f_i(w_{t,\tau}^i))\right\|^2\notag\\&+\frac{ \zeta_t^2L}{2n^2}\mathbb{E}\left\|\sum_{i \in Z_t} \sum_{\tau=0}^{K-1} \nabla f_i(w_{t,\tau}^i)\right\|^2 \notag\\ \overset{(b)}{\leq}& \frac{ K\zeta_t^2 L\sigma_l^2}
{2n}  + \frac{\zeta_t^2 L}{2n^2}\mathbb{E}\left\|\sum_{i \in Z_t} \sum_{\tau=0}^{K-1} \nabla f_i(w_{t,\tau}^i)\right\|^2,
\end{align}
where inequality $(a)$ is guaranteed by Assumption \ref{assumption:2}, and $(b)$ is established through both Assumption \ref{assumption:2} and the Cauchy-Schwarz inequality. By defining \(x_i := \sum_{\tau=0}^{K-1} \nabla f_i(w_{t,\tau}^i)\), we have 
\(
\mathbb{E}\left\|\sum_{i \in Z_t} x_i\right\|^2 = \mathbb{E}\left\|\sum_{i \in Z_t} \sum_{\tau=0}^{K-1} \nabla f_i(w_{t,\tau}^i)\right\|^2,
\)
and we derive the following expression based on the properties of random sampling without replacement for \(Z_t\).
\begin{align}
\label{eq:27}
& \mathbb{E}\|\sum_{i \in Z_t} x_i\|^2 = \mathbb{E}[\sum_{i \in Z_t} \|x_i\|^2 + \sum_{\binom{i \neq j}{i,j \in Z_t}} \langle x_i, x_j \rangle] \notag \\ 
& = \mathbb{E}[\sum_{i \in Z_t} \|x_i\|^2] + \mathbb{E}[\sum_{\binom{i \neq j}{i,j \in Z_t}} \langle x_i, x_j \rangle] \notag \\ 
& = \frac{n}{N} \sum_{i \in [N]} \mathbb{E}[ \|x_i\|^2] + \frac{n(n-1)}{N(N-1)} \mathbb{E}[\sum_{i \neq j} \langle x_i, x_j \rangle] \notag \\ 
& = \frac{n}{N} \sum_{i \in [N]} \mathbb{E}[\|x_i\|^2] + \frac{n(n-1)}{N(N-1)} (\mathbb{E}\|\sum_{i \in [N]} x_i\|^2 - \sum_{i \in [N]} \mathbb{E}[\|x_i\|^2]) \notag \\ 
& = [\frac{n}{N} - \frac{n(n-1)}{N(N-1)}] \sum_{i \in [N]}  \mathbb{E}[\|x_i\|^2] + \frac{n(n-1)}{N(N-1)} \mathbb{E}\|\sum_{i \in [N]} x_i\|^2.
\end{align}
Next, we analyze the upper bound of $\sum_{i \in [N]} \mathbb{E} [\|x_i\|^2]$.
\begin{align}
\label{eq:28}
&\sum_{i \in [N]} \mathbb{E} [\|x_i\|^2] \\ \notag =& \sum_{i \in [N]} \mathbb{E} [\|\sum_{\tau=0}^{K-1} [\nabla f_i(w_{t,\tau}^i)-\nabla f_i(w_t) + \nabla f_i(w_t) \\ \notag &-\nabla f(w_t) + \nabla f(w_t) ]\|^2] \\ \notag \leq & 3KL^2 \sum_{i \in [N]}\sum_{\tau=0}^{K-1}\mathbb{E}\|w_{t,\tau}^i-w_t\|^2 + 3K^2N \sigma_g^2 + 3K^2N \mathbb{E}\|\nabla f(w_t)\|^2 \\ \notag \leq& 9 N K^3 L^2 \zeta_t^2(\sigma_l^2+6K\sigma_g^2) + (54 N K^4 \zeta_t^2 L^2 + 3K^2N) \mathbb{E}\| \nabla f(w_t)\|^2 \\ \notag  &+ 3K^2N\sigma_g^2.
\end{align}
Substitute (\ref{eq:28}), (\ref{eq:27}) into  (\ref{aeq:6}), we can obtain
\begin{align}
\label{A_2}
A_2 & \leq   \frac{ K\zeta_t^2 L\sigma_l^2}{2n}  +   \frac{\zeta_t^2 L}{2n^2}  ([\frac{n}{N} - \frac{n(n-1)}{N(N-1)}] [9 N K^3 L^2 \zeta_t^2 \times \\ \notag &(\sigma_l^2+6K\sigma_g^2) + (54 N K^4 \zeta_t^2 L^2 + 3K^2N) \mathbb{E}\| \nabla f(w_t)\|^2 \\ \notag  &+ 3K^2N\sigma_g^2] + \frac{n(n-1)}{N(N-1)} \mathbb{E}\|\sum_{i \in [N]}x_i\|^2) 
\end{align}

Substitute (\ref{A_2}), (\ref{aeq:5}) into  (\ref{aeq:1}), we can obtain
\begin{align}
\label{aeq:7}
&\mathbb{E}f(\hat{w}_{t+1}) \leq \mathbb{E}f(w_t)-K\zeta_t(\frac{1}{2}-9K^2\zeta_t^2L^2\notag\\&- \frac{\zeta_t L (N-n)}{2n(N-1)}(54K^3\zeta_t^2L^2+3K)) \mathbb{E}\left\|\nabla f(w_t)\right\|^2\notag\\&+(\frac{\zeta_t^2L(n-1)}{2nN(N-1)}- \frac{\zeta_t}{2N^2K})\mathbb{E}\left\|\sum_{i \in [N]} x_i\right\|^2\notag\\&+  (\frac{3}{2} K^2 \zeta_t^3 L^2 + \frac{(N-n)}{2n(N-1)}(9K^3L^3\zeta_t^4))(\sigma_l^2+6K\sigma_g^2)  \notag \\ &+ \frac{ K\zeta_t^2 L\sigma_l^2}{2n}+ \frac{(N-n)}{2n(N-1)} 3K^2L\zeta_t^2\sigma_g^2 \notag \\ \leq & \mathbb{E}f(w_t)-C\zeta_tK \mathbb{E}\|\nabla f(w_t)\|^2 + \frac{(N-n)}{2n(N-1)} 3K^2L\zeta_t^2\sigma_g^2 \notag \\ &+ (\frac{3}{2} K^2 \zeta_t^3 L^2 + \frac{(N-n)}{2n(N-1)}(9K^3L^3\zeta_t^4))(\sigma_l^2+6K\sigma_g^2) \notag \\ &+ \frac{ K\zeta_t^2 L\sigma_l^2}{2n} .
\end{align}
The last inequality follows from \(\frac{\zeta_t^2L(n-1)}{2nN(N-1)} - \frac{\zeta_t}{2N^2K} \leq 0\) if \(\zeta_tKL \leq \frac{n(N-1)}{N(n-1)}\), and from the existence of a constant \(C > 0\) such that  
\(
\frac{1}{2} - 9K^2\zeta_t^2L^2 - \frac{\zeta_t L (N-n)}{2n(N-1)}\left(54K^3\zeta_t^2L^2 + 3K\right) \geq C.
\)    
\end{proof}
\subsection{Proof of Lemma \ref{lem:2}}
\begin{proof}
Next, we delve into another aspect of server-side training. Due to the Assumption \ref{assumption:1}, we have
\begin{align}
\label{aeq:8}
\mathbb{E}f(w_{t+1}) =& \mathbb{E}f\left(\hat{w}_{t+1}-\sum_{e=0}^{E-1}\eta_t \tilde{\nabla} f_\mathcal A (\hat{w}_{t+1,e})\right) \notag\\ \leq &\mathbb{E}f(\hat{w}_{t+1}) + \underbrace{\mathbb{E}\langle\nabla f(\hat{w}_{t+1}), -\sum_{e=0}^{E-1}\eta_t\Tilde{\nabla}f_\mathcal{A}(\hat{w}_{t+1,e})\rangle}_{A_4}
\notag\\&+ \underbrace{\frac{L}{2}\mathbb{E}\left\|
\eta_t \sum_{e=0}^{E-1}\Tilde{\nabla} f_\mathcal{A}(\hat{w}_{t+1,e})\right\|^2.}_{A_5}
\end{align}

Now, we analyze the upper bound of $A_4$.
\begin{align}
\label{aeq:9}
A_4=& \mathbb{E}\langle \nabla f(\hat{w}_{t+1}),-\sum_{e=0}^{E-1}\eta_t\Tilde{\nabla} f_\mathcal{A}(\hat{w}_{t+1,e})\rangle
\notag \\ \overset{(a)}{=}& -\sum_{e=0}^{E-1}\eta_t\mathbb{E}\langle \nabla f(\hat{w}_{t+1}),\nabla f_ \mathcal{A}
(\hat{w}_{t+1,e})\rangle\notag\\ \overset{(b)}=&-\sum_{e=0}^{E-1}\eta_t\mathbb{E}\langle \nabla f(\hat{w}_{t+1}),\nabla f(\hat{w}_{t+1,e})\rangle\notag\\
\overset{(c)}{=}&-\sum_{e=0}^{E-1}\eta_t \mathbb{E}(\frac{1}{2}\left\|\nabla f(\hat{w}_{t+1})\right\|^2+ \frac{1}{2}\left\|\nabla f(\hat{w}_{t+1,e})\right\|^2 \notag\\&- \frac{1}{2}\left\|\nabla f(\hat{w}_{t+1})-\nabla f(\hat{w}_{t+1,e})\right\|^2)\notag\\ \overset{(d)}{\leq}& - \sum_{e=0}^{E-1} \eta_t\mathbb{E}(\frac{1}{2}\left\|\nabla f(\hat{w}_{t+1})\right\|^2+\frac{1}{2}\left\|\nabla f(\hat{w}_{t+1,e})\right\|^2 \notag\\&-\frac{L^2}{2}\left\|\hat{w}_{t+1}-\hat{w}_{t+1,e}\right\|^2),
\end{align}
where $(a)$ holds due to Assumption \ref{assumption:2}, $(b)$ is ensured by Assumption \ref{assumption:4}, $(c)$ is established through the equation $\langle a, b \rangle = \frac{1}{2}\|a\|^2 + \frac{1}{2}\|b\|^2 - \frac{1}{2}\|a-b\|^2$ for any vectors $a$ and $b$, and $(d)$ is supported by Assumption \ref{assumption:1}.
\begin{align}
\label{aeq:10}
&\mathbb{E}\left\|\hat{w}_{t+1} - \hat{w}_{t+1,e}\right\|^2
\notag\\=&\mathbb{E}\left\|\eta_t\sum_{p=0}^{e-1}\Tilde{\nabla}f_\mathcal{A}(\hat{w}_{t+1,p})\right\|^2 \notag\\\ \overset{(a)}{\leq}& \eta_t^2 e \sum_{p=0}^{e-1}\mathbb{E}\left\|\Tilde{\nabla} f_\mathcal{A}(\hat{w}_{t+1,p})\right\|^2 
\notag\\\leq& \eta_t^2 e \sum_{p=0}^{e-1}\mathbb{E}\left\|\Tilde{\nabla} f _ \mathcal{A}(\hat{w}_{t+1,p})-\nabla f_\mathcal{A}(\hat{w}_{t+1,p})+\nabla f_ \mathcal{A}(\hat{w}_{t+1,p})\right\|^2
\notag\\ \overset{(b)}{=}&\eta_t^2e\sum_{p=0}^{e-1}\mathbb{E}(\left\|\nabla f_\mathcal{A}(\hat{w}_{t+1,p})\right\|^2+\left\|\Tilde{\nabla} f_ \mathcal{A}(\hat{w}_{t+1,p})-\nabla f_\mathcal{A}(\hat{w}_{t+1,p})\right\|^2)\notag\\ \overset{(c)}{=}&\eta_t^2e\sum_{p=0}^{e-1}\mathbb{E}\left\|\nabla f_\mathcal{A}(\hat{w}_{t+1,p})\right\|^2 + \eta_t^2e^2\sigma_l^2,
\end{align}
where $(a)$ holds due to the Cauchy-Schwarz inequality and $(b),(c)$ holds due to the Assumption \ref{assumption:2} and  \ref{assumption:3}.
We substitute (\ref{aeq:10}) into  (\ref{aeq:9}) to get
\begin{align}
\label{aeq:11}
A_4 \leq& - \sum_{e=0}^{E-1} \eta_t\mathbb{E}\Big(\frac{1}{2}\left\|\nabla f(\hat{w}_{t+1})\right\|^2+\frac{1}{2}\left\|\nabla f_\mathcal{A}(\hat{w}_{t+1,e})\right\|^2 \notag\\ &- \frac{L^2}{2}(\eta_t^2e\sum_{p=0}^{e-1}\left\|\nabla f_\mathcal{A}(\hat{w}_{t+1,p})\right\|^2 + \eta_t^2e^2\sigma_l^2)\Big)
\notag\\=& -\frac{\eta_tE}{2}\mathbb{E}\left\|\nabla f(\hat{w}_{t+1})\right\|^2-\frac{\eta_t}{2}\sum_{e=0}^{E-1}\mathbb{E}\left\|\nabla f_\mathcal{A} (\hat{w}_{t+1,e})\right\|^2 \notag\\&+ \frac{\eta_t^3L^2}{2}\sum_{e=0}^{E-1}e\sum_{p=0}^{e-1}\mathbb{E}\left\|\nabla f_\mathcal{A}(\hat{w}_{t+1,p})\right\|^2 + \frac{L^2\eta_t^3\sigma_l^2}{2} \sum_{e=0}^{E-1} e^2
\notag\\\leq& - \frac{\eta_t E}{2} \mathbb{E}\left\|\nabla f(\hat{w}_{t+1})\right\|^2- \frac{\eta_t}{2}\sum_{e=0}^{E-1}\mathbb{E}\left\|\nabla f_\mathcal{A}(\hat{w}_{t+1,e})\right\|^2
\notag\\&+ \frac{\eta_t^3 L^2 E^2}{4}\sum_{e=0}^{E-1}\mathbb{E}\left\|\nabla f_\mathcal{A}(\hat{w}_{t+1,e})\right\|^2 \notag\\&+ \frac{L^2\eta_t^3\sigma_l^2 E(E-1)(2E-1)}{12}.
\end{align}
Let's bound $A_5$ next
\begin{align}
\label{aeq:12}
A_5 =&\frac{L}{2}\mathbb{E}\|
\eta_t \sum_{e=0}^{E-1}\Tilde{\nabla} f_\mathcal{A}(\hat{w}_{t+1,e})\|^2 \notag\\ \overset{(a)}{\leq}& \frac{L\eta_t^2E}{2}\sum_{e=0}^{E-1}\mathbb{E}\left\|\Tilde{\nabla} f _\mathcal{A}(\hat{w}_{t+1,e})\right\|^2
\notag\\=&\frac{L \eta_t^2E}{2}\sum_{e=0}^{E-1}\mathbb{E}\|(\Tilde{\nabla} f_\mathcal{A}(\hat{w}_{t+1,e})-\nabla f_\mathcal{A}(\hat{w}_{t+1,e})\notag\\&+\nabla f_\mathcal{A}(\hat{w}_{t+1,e}))\|^2
\notag\\ \overset{(b)}{=}&\frac{L\eta_t^2E}{2}\sum_{e=0}^{E-1}\mathbb{E}\left\|\Tilde{\nabla} f _\mathcal{A}(\hat{w}_{t+1,e})-\nabla f_\mathcal{A}(\hat{w}_{t+1,e})\right\|^2 \notag\\&+ \frac{L\eta_t^2E}{2}\sum_{e=0}^{E-1}\mathbb{E}\left\|\nabla f_\mathcal{A}(\hat{w}_{t+1,e})\right\|^2
\notag\\ \overset{(c)}{\leq}& \frac{L\eta_t^2E^2\sigma_l^2}{2}+\frac{L\eta_t^2E}{2}\sum_{e=0}^{E-1}\mathbb{E}\left\|\nabla f_\mathcal{A}(\hat{w}_{t+1,e})\right\|^2,
\end{align}
where $(a)$ is established because of the Cauchy-Schwartz inequality, and $(b),(c)$ is established because of the Assumption \ref{assumption:2} and \ref{assumption:3}.

Substituting (\ref{aeq:11}), (\ref{aeq:12}) into  (\ref{aeq:8}), we can get
\begin{align}
\label{aeq:13}
\mathbb{E}f(w_{t+1}) \leq & \mathbb{E}f(\hat{w}_{t+1}) - \frac{\eta_t E}{2} \mathbb{E}\left\|\nabla f(\hat{w}_{t+1})\right\|^2
 \notag\\ &+\frac{L^2\eta_t^3 E(E-1)(2E-1)\sigma_l^2}{12}+ \frac{L\eta_t^2E^2\sigma_l^2}{2}\notag\\-&(\frac{\eta_t}{2}-\frac{L^2\eta_t^3E^2}{4}-\frac{L\eta_t^2E}{2})\sum_{e=0}^{E-1}\mathbb{E}\left\|\nabla f_\mathcal{A}(\hat{w}_{t+1,e})\right\|^2
\notag\\\leq& \mathbb E f(\hat{w}_{t+1})-\frac{\eta_tE}{2}\mathbb{E}\left\|\nabla f(\hat{w}_{t+1})\right\|^2+\frac{L\eta_t^2E^2\sigma_l^2}{2}\notag\\&+\frac{L^2\eta_t^3 E(E-1)(2E-1)\sigma_l^2}{12},
\end{align}
where the last inequality holds because 
\(\frac{\eta_t}{2} - \frac{L^2\eta_t^3E^2}{4} - \frac{L\eta_t^2E}{2} \geq 0\),
which follows from \(\eta_t = \frac{1}{LE\sqrt{T}}\) and holds due to \(T \geq \frac{2+\sqrt{3}}{2}\).
\end{proof}
\end{appendices}

\bibliographystyle{IEEEtran}
\bibliography{references}

\end{document}